\documentclass[10pt]{wlscirep}
\usepackage[utf8]{inputenc}
\usepackage[T1]{fontenc}
\usepackage{float}
\usepackage{longtable}
\usepackage{todonotes}
\usepackage{booktabs}
\usepackage{array}
\usepackage{graphicx}
\newcommand{\format}[1]{\emph{\textcolor{blue}{#1}}}
\title{Information Anxiety in Large Language Models}

\author[1]{Prasoon Bajpai}
\author[2]{Sarah Masud}
\author[3,4,*]{Tanmoy Chakraborty}
\affil[1]{Department of Mathematics, Indian Institute of Technology, Delhi, New Delhi, India}
\affil[2]{ Department of Computer Science, University of Copenhagen, Denmark}
\affil[3]{Department of Electrical Engineering, Indian Institute of Technology, Delhi, New Delhi, India}
\affil[4]{Yardi School of Artificial Intelligence, Indian Institute of Technology, Delhi, New Delhi, India}
\affil[*]{Corresponding author: tanchak@iitd.ac.in}

\begin{abstract}

Large Language Models (LLMs) have demonstrated strong performance as knowledge repositories, enabling models to understand user queries and generate accurate and context-aware responses. Extensive evaluation setups have corroborated the positive correlation between the retrieval capability of LLMs and the frequency of entities in their pretraining corpus. We take the investigation further by conducting a comprehensive analysis of the internal reasoning and retrieval mechanisms of LLMs. Our work focuses on three critical dimensions --- the impact of entity popularity, the models' sensitivity to lexical variations in query formulation, and the progression of hidden state representations across LLM layers. Our preliminary findings reveal that popular questions facilitate early convergence of internal states toward the correct answer. However, as the popularity of a query increases, retrieved attributes across lexical variations become increasingly dissimilar and less accurate. Interestingly, we find that LLMs struggle to disentangle facts, grounded in distinct relations, from their parametric memory when dealing with highly popular subjects. Through a case study, we explore these latent strains within LLMs when processing highly popular queries, a phenomenon we term \textit{information anxiety}. The emergence of information anxiety in LLMs underscores the adversarial injection in the form of linguistic variations and calls for a more holistic evaluation of frequently occurring entities.
\end{abstract}
\begin{document}
\flushbottom
\maketitle
\section*{Introduction}
The ability to perform effective question answering (QA) serves as the backbone of modern artificial intelligence systems\cite{zhao2024surveylargelanguagemodels, shabbir2018artificialintelligencerolenear}. Research in the field of question answering\cite{zhang-etal-2023-survey-efficient} dates back to conversational systems, like ELIZA\cite{10.1145/365153.365168}, or work in the field of psychology on how people interpret and understand language\cite{doi:10.1111/j.1467-9280.1990.tb00059.x, Casasanto2004HowDA} and retrieve information from their memory\cite{Nagel2024}. In recent years, Large Language Models (LLMs) have achieved state-of-the-art on popular QA benchmarks such as SQuAD\cite{rajpurkar2016squad100000questionsmachine}, TriviaQA\cite{joshi-etal-2017-triviaqa}, and Natural Questions\cite{kamalloo-etal-2023-evaluating}. Several studies have identified a positive correlation between the frequency of information in the pretraining corpora of LLMs and the extent of their memorization\cite{RICHEY2019173,carlini2023quantifyingmemorizationneurallanguage,kandpal2023largelanguagemodelsstruggle}. 
\begin{figure}[ht!]
    \centering
    \includegraphics[width=\linewidth]{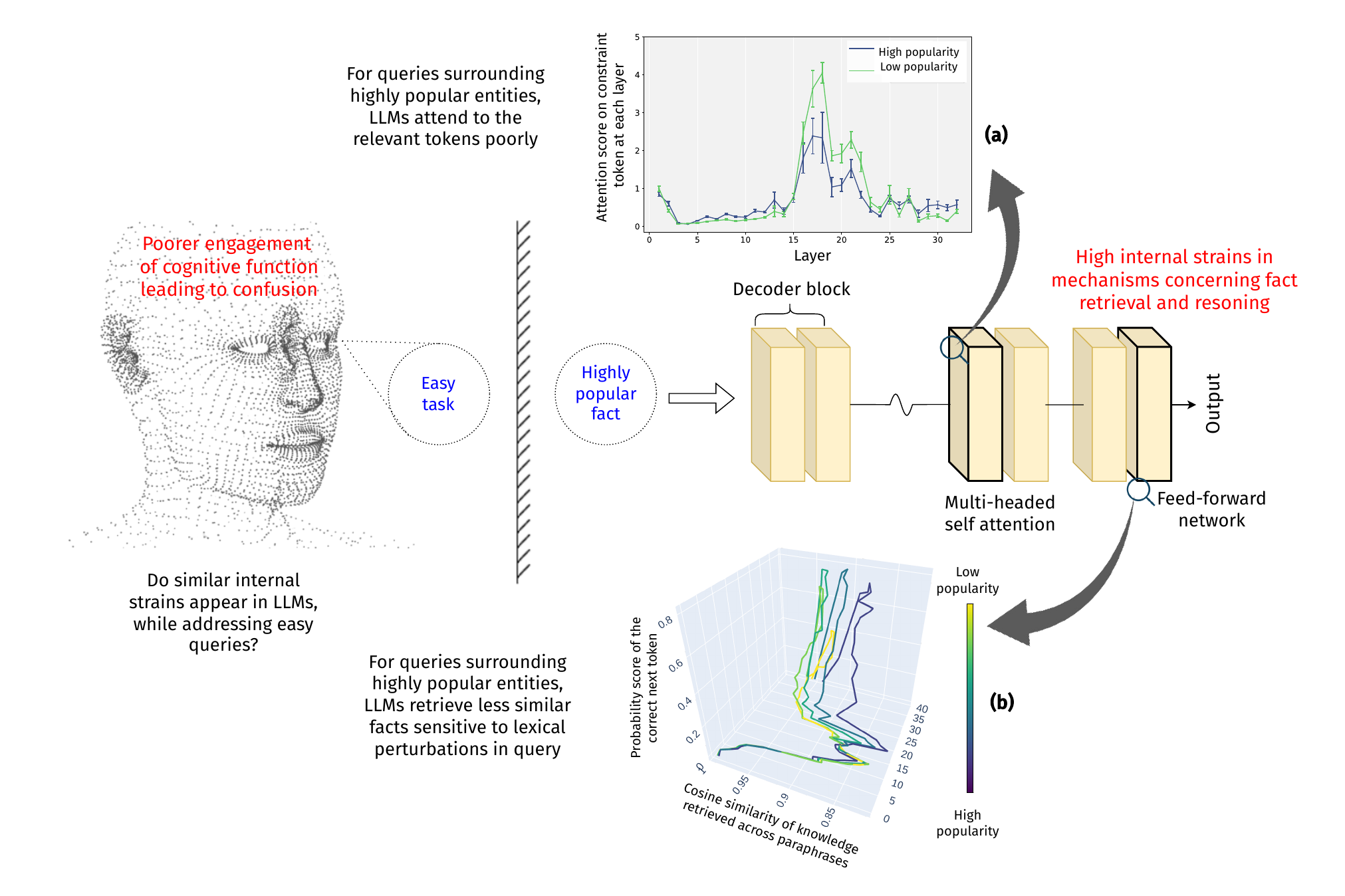}
    \caption{\textbf{Overview of our findings}. The figure demonstrates our proposed experimental setup, where we prompt models from the Llama-2 family with factual questions from \text{PopQA} in an in-context learning setting. We intercept the hidden states after each decoder block and invert it to obtain the next token probability distribution. We find that for questions based on highly popular entities, predicted probability distributions converge rapidly to the target probability distributions. \textbf{(a)} The figure demonstrates the difference in the attention scores given to a set of specific tokens based on the popularity of questions. There is a higher attention score provided, in case of lower popularity questions. \textbf{(b)} The figure demonstrates the similarity of the facts retrieved across different lexical variations of the same question, depending on the questions’s popularity. A higher dissimilarity for the lexical variations of the same question, is observed given the popularity of the question is high. Our findings expose critical internal strains in LLMs while addressing highly prevalent information, a phenomena we term as \textit{information anxiety}.}
    \label{fig:intro_flow}
\end{figure}
\begin{figure}[ht]
    \centering
    \includegraphics[width=\linewidth]{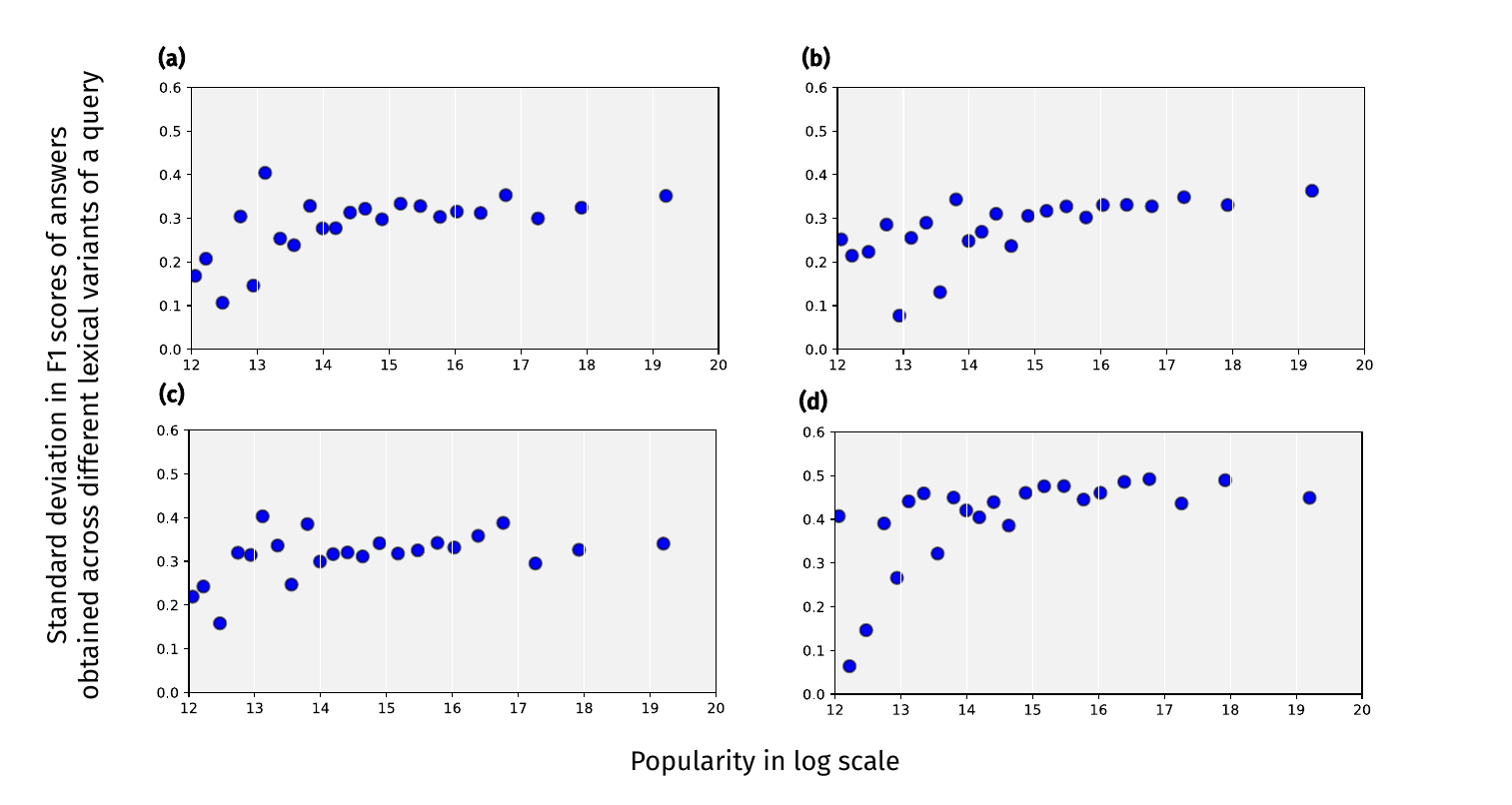}
    \caption{\textbf{Variety in responses across different scales of popularity}. For each question, we capture the variety in responses by calculating the F1 scores of answers across the different lexical variants. We can see, for the `Screenwriter’ relation across four models, that there is an increase in the variety of responses with an increase in the popularity of questions, suggesting the sensitivity of responses to the lexical structure of queries based on highly popular entities. The figure shows results for following models -  \textbf{(a)} Llama-2-7B, \textbf{(b)} Llama-2-7B-chat, \textbf{(c)} Llama-2-13B, and \textbf{(d)} Llama-2-13B-chat.}
    \label{fig:introduction_variety_screenwriter}
\end{figure}

However, these studies only observe the effect of popularity at a macroscopic level by focusing on performance-based metrics in QA tasks. Given the critical role of QA systems in today's AI space, it is imperative to evaluate the effect of ``popularity" on the internal computational of an LLM. Specifically, the aim is to inquire: \textit{Do queries concerning widely recognized entities introduce ambiguity or confusion in LLMs}? Our investigation is primarily inspired by psycholinguistic studies\cite{abbas_taskcomplexity,amini_taskfamiliarityordering,RICHEY2019173}, which indicate that humans tend to perform better on tasks perceived as more complex, suggesting that easier tasks may not engage cognitive resources effectively, leading to poorer performance. We seek to find similar patterns in LLMs by investigating the internal dissonance faced by LLMs when responding to queries of varying popularity. To incorporate this phenomenon into our study, we introduce the term \textit{information anxiety}, defined as the latent incongruity and strain on the knowledge-retrieval mechanisms of LLMs as they generate responses to queries.

To capture the effect of popularity on the variety of responses in a QA setup, we employ PopQA \cite{mallen-etal-2023-trust}, which is an entity-centric open-domain dataset. In addition to QA pairs, the dataset contains the entity and their relations in the form of a triplet $(s, r, o)$, where $s$, $r$, and $o$ represent the subject, relation, and object (target response), respectively \cite{DBLP:journals/corr/abs-1909-01066}. For example, the triplet \text{$\langle$Ireland, capital, Dublin$\rangle$} corresponds to the query ``\textit{What is the capital of Ireland?}'' The dataset also enlists Wikipedia page hits as a proxy for the popularity of subjects and objects. In our setup (as shown in Figure \ref{fig:intro_flow}), the popularity of a query/question refers to the mean popularity of the subject and object entities in the question. 

To observe the effect of popularity on the variety of final responses, we create the following preliminary experiment setup. For each query $Q_{i}$, we create $p = 10$ lexical variations $\{P_{i}^{j}:  1 \leq j \leq p\}$, while retaining its semantical meaning. The evaluation is carried out via a lexical matching of predicted object tokens to a set of golden tokens using F1 scores. For query $Q_{i}$, we capture a variety in responses, \texttt{Var}($i$) as the standard deviation in the F1 score for the answers across the $p$ lexical variants. Figure \ref{fig:introduction_variety_screenwriter} demonstrates that the variety in responses. \texttt{Var}($\cdot$) increases with the logarithm of the popularity of underlying questions. Such increased variety in responses raises questions about robustness of LLMs as QA models, especially while addressing highly prevalent information. This motivates us to create a controlled set of experiments to unveil some microscopic patterns in the internal mechanisms of LLMs, while dealing with highly popular questions.

We use models of various scales from the open-source Llama-2 family\cite{touvron2023llama2openfoundation} (Llama-2-7B, Llama-2-7B-chat, Llama-2-13B, Llama-2-13B-chat and Llama-2-70B) under few-shot prompting to test our hypothesis. Using the setup in Figure \ref{fig:intro_flow}, we examine information anxiety via the following research questions (RQs):
\vspace{-0.2cm}
\begin{itemize}
    \itemsep-0.04cm
    \item\textbf{RQ1:} What is the effect of popularity on the internal hidden states of LLMs?
    \item \textbf{RQ2:} Do LLMs attend to specific portions of a question differently depending on the question's popularity?
    \item \textbf{RQ3:} What is the effect of lexical variations and popularity of the questions on the knowledge retrieved from an LLM's parametric memory?
    \item \textbf{RQ4:} Do LLMs struggle to segregate internal knowledge of similar relations when dealing with popular questions?
\end{itemize}

Firstly, at a macroscopic level, in line with the status quo, we observe that for highly popular questions, the predicted probability distribution for the next token converges much more rapidly towards the target probability distribution than for less popular questions (\text{RQ1}). However, looking at fine-grained attention weights, we find a lower attention score being accumulated over a relevant set of tokens when addressing questions with higher popularity. The relevant set of tokens includes the subject of the question and the words describing the relation. Our findings suggest that when understanding a question containing a highly popular subject, LLMs tend to overlook essential parts of the query (RQ2). It is akin to the concept of muscle memory in humans.
Consequently, it propels the examination of information anxiety that emerges when lexical variations of a popular query are introduced. Albeit a high level of confidence in answering popular queries (RQ3), with increasing popularity, a higher level of dissimilar tokens are retrieved for variants of the same query. This inconsistency of retrieved tokens under a high probability environment signals an adverse effect of popularity on the internal retrieval mechanisms of LLMs. This is further corroborated by the observation that LLMs fail to segregate knowledge about highly popular distinct relations (RQ4). Observations from RQs 2-4 expose critical internal strains in LLMs while dealing with highly frequent information. The prevalence of \text{information anxiety} raises significant concerns over the adversary that lexical variations can trigger in an LLM. 
 
\section*{Results}
\begin{figure}[!t]
    \centering
    \includegraphics[width=0.75\linewidth]{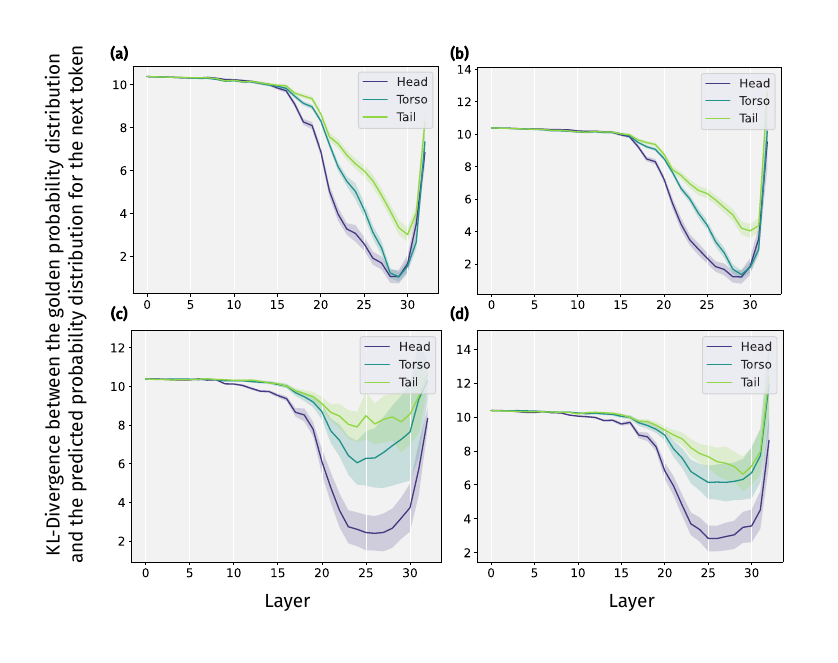}
    \caption{\textbf{Convergence of the predicted probability distribution}. The figure shows the KL-Divergence between the next token predicted probability distribution and the golden probability distribution for questions of \texttt{capital} and \texttt{genre} relations in \texttt{PopQA}. For each batch of questions, the solid line denotes the average of the distances between the predicted and the golden distributions across all questions in a batch (\texttt{Head}, \texttt{Torso} or \texttt{Tail}). We also account for the effect of lexical variations on the next token probability distribution. The shaded regions denote the average impact of such variations (captured using deviations in the KL-Divergence distances) across all questions in a batch. We observe the predicted probability distribution associated with highly popular questions approaches the golden probability distribution earlier as compared to that associated with comparatively less popular questions. The figure shows for following models and relation combination --- \textbf{(a)} Llama-2-7B + \texttt{capital}, \textbf{(b)} Llama-2-7B-chat + \texttt{capital}, \textbf{(c)} Llama-2-7B + \texttt{genre}, and \textbf{(d)} Llama-2-7B-chat + \texttt{genre}. All remaining results can be found in \textit{Supplementary}, Section 2.1.} 
    \label{fig:result_distance-hidden_capital_genre}
\end{figure}
To reiterate, we use PopQA\cite{mallen-etal-2023-trust}, an entity-centric open-domain QA dataset containing 14k QA pairs  Each entry of PopQA is grounded on entities: subject ($s$) and object ($s$), related through a relation ($r$)  The popularity of a question is measured by the average of its underlying subject's and object's popularity (see \textit{Methods} for the detailed dataset description)  For each query in PopQA, we create ten different lexical perturbations while maintaining the underlying semantic sense  Exact templates for lexical variation of each relation in PopQA can be found in \textit{Supplementary}, Section S1.1.  Using PopQA, we design experiments to observe the effect of query popularity on the internal retrieval and reasoning mechanisms of LLMs (see \textit{Methods} for a brief overview of the Transformer model)  We find that LLMs poorly attend to relevant sections of queries that pertain to highly popular entities  Moreover, we observe a marked dissimilarity in the facts retrieved for the same query when prompted with lexical variations. The poor attention to relevant tokens and the dissimilarity of retrieved facts raise questions about the robustness of LLMs as QA models. The observations are in contrast with the preconceived notion of performing really well externally while addressing highly prevalent information. We report a marked internal stress in core components of LLMs when dealing with highly prevalent information, a phenomenon we term information anxiety in LLMs.

\subsection*{Effect of Popularity on Convergence of the Predicted Probability Distribution}

Let \texttt{Head} (highly popular), \texttt{Torso} (medium popular), and \texttt{Tail} (less popular) denote sets of 50 questions each, in decreasing order of average popularity. To find how the popularity of a question affects the evolution of hidden states inside an LLM, we compare via KL-Divergence the predicted probability distribution for the next token prediction with a probability distribution having its entire mass on the correct target token (we call it the golden probability distribution). To obtain the predicted probability distribution for the next token, we invert the hidden states after each decoder block (see \textit{Methods} for a more detailed description of the experimental setup). The exhaustive list of results across all models and relations can be found in \textit{Supplementary}, Figure S1. Succinctly, the results in Figure \ref{fig:result_distance-hidden_capital_genre} highlight that the distance between the predicted and the golden probability distribution decreases much more rapidly for the batch of highly popular questions (Head), as compared to the batch of less popular questions (Torso \& Tail). Further, across most classes of popularity and models, the distance between the predicted probability distribution and golden probability distribution increases in the final layers before producing output tokens. This suggests that the popularity of a question has a distinct influence on the hidden states of an LLM. Moreover, at a particular layer, the effect of linguistic variations on the distance between the next token probability distribution and the golden probability distribution increases as the next token distribution converges towards the golden distribution. In the rest of the experiments, we isolate the effect of a question's popularity on internal reasoning as well as the fact retrieval structures of LLMs.

\subsection*{Effect of Popularity on LLM Reasoning Behaviour}
\begin{figure}[!t]
    \centering
    \includegraphics[width=0.8\linewidth]{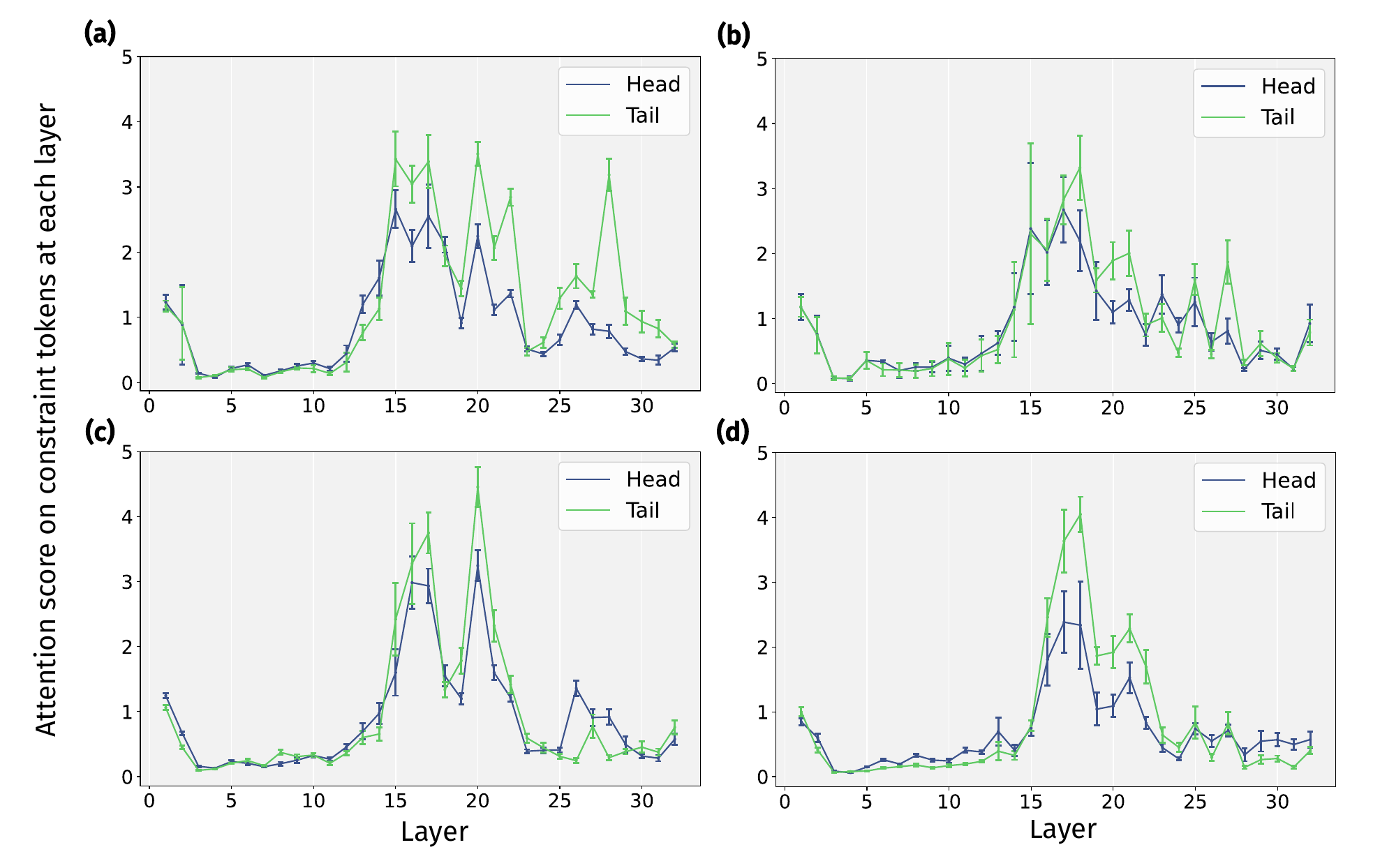}
    \caption{\textbf{Effect of popularity on LLM reasoning behaviour}. For each lexical variant of a question, we calculate a measure of attention scores over each constrain token. For each question, we take a mean of these attention scores across all its lexical variants and show it with solid lines for each layer under study. We also take the standard deviation across all of the lexical variants of a question and show it using error bars for each layer. \texttt{Head} denotes the batch of highly popular questions while \texttt{Tail} denotes the batch of less popular questions. The attention score, as well as its sensitivity to lexical variation (observed from the standard deviation), appears to be highest in the middle layers of processing. Furthermore, we observe that between \texttt{Head} and \texttt{Tail}, a lower attention score is associated with the batch of highly popular questions. The figure shows the results for following model + relation combination --- \textbf{(a)} Llama-2-7B + {\tt Capital}, \textbf{(b)} Llama-2-7B + {\tt Occupation}, \textbf{(c)} Llama-2-7B-chat + {\tt Capital}, \textbf{(d)} Llama-2-7B-chat + {\tt Occupation}. All results can be found in \textit{Supplementary}, Section 2.2.}
    \label{fig:result_attention_captial_occupation}
\end{figure}
For each lexical variation of a question, we define a set of constraint tokens based on relations and subjects modified in the variant. For a linguistic variant, we record the attention score for a constraint token averaged across all attention heads. The overall impact registered by a lexical variant is based on the maximum score of all its constraint tokens. We then measure the central tendencies of these scores across the variants of a query. We elaborate on the technical details of the experimental setup in \textit{Methods}. Contrary to popular belief that popular queries are easy to operate, Figure \ref{fig:result_attention_captial_occupation} highlights that for less popular questions, LLMs address the constraint tokens better than questions of higher popularity. Consequently, high-popularity queries show signs of information anxiety, where depending on the supposed memorized knowledge, the LLM pays less attention to the constrained tokens. We hypothesize that this can cause inconsistency in factual recall in the subsequent feed-forward networks and examine the same in our next RQ. The exhaustive list of results is available in \textit{Supplementary}, Figure S2. 


\begin{figure}[ht]
    \centering
    \includegraphics[width=0.9\linewidth]{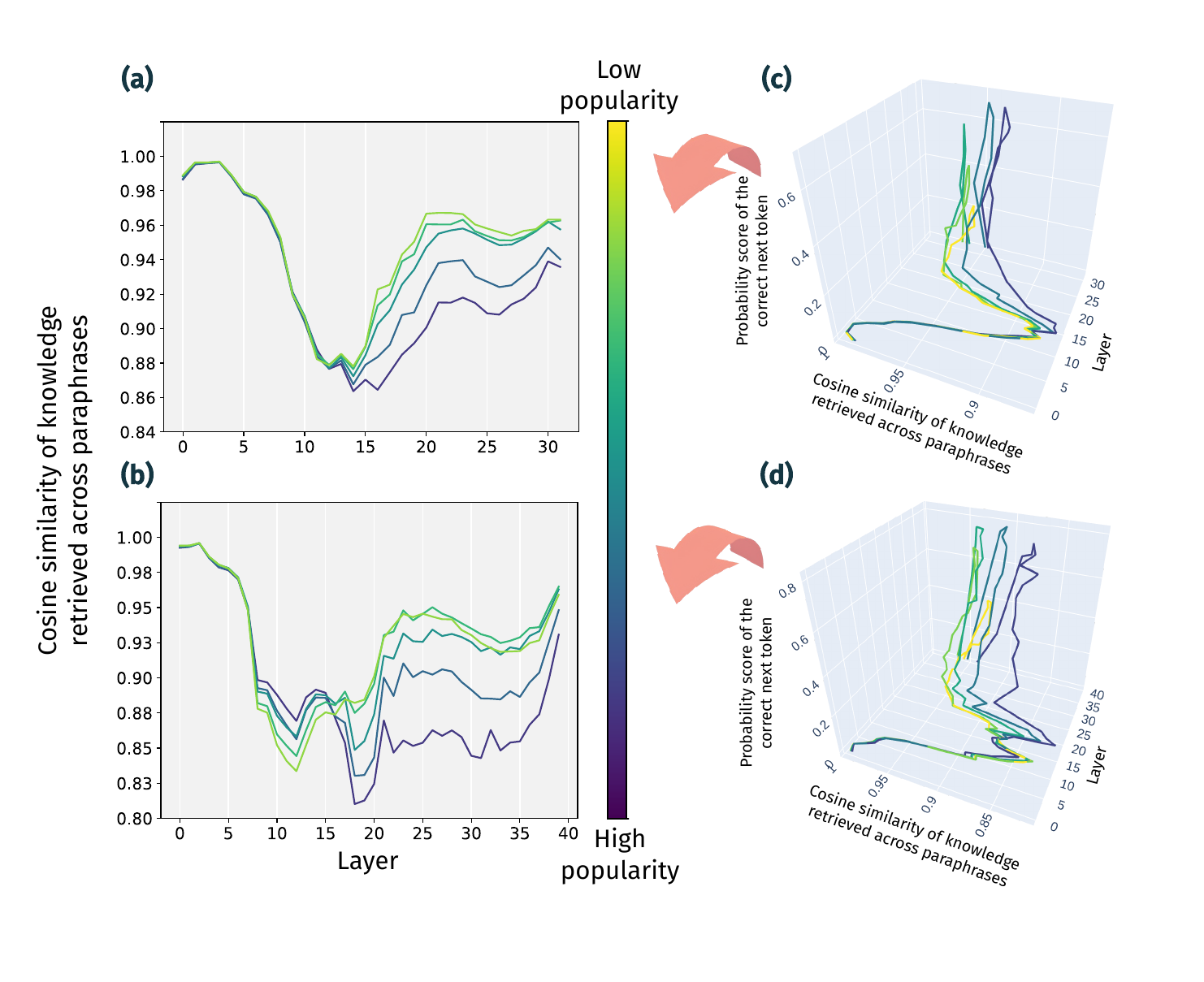}
    \caption{\textbf{Effect of popularity on LLM fact-retrieval behaviour}. (a) and (b) demonstrate the similarity between the facts retrieved corresponding to different lexical variations of each question. We observe a lower similarity between the facts retrieved across different lexical variations of a prevalent question. (c) and (d) add an extra axis showing the batch-wise representation of the target token probability score finally sampled from the predicted probability distribution at each layer. While there is an inconsistency in the facts retrieved, LLMs stay confident in their prediction despite sensitivity towards lexical variations of highly popular questions. The figure shows results for following models -- \textbf{(a), (c)} Llama-2-7B + {\tt Capital}, \textbf{(b), (d)} Llama-2-13B + {\tt Capital}. Results for the other LLMs can be found in $Supplementary$, Section 2.3.}
    \label{fig:result_neural_capital}
\end{figure}

\subsection*{Effect of Popularity on LLM Fact Retrieval Behaviour}

Motivated by our previous result, we assess how the difference in the lexical structure and popularity of a question affects the fact retrieval behaviour in LLMs. At each layer of processing, we intercept the upward projection of the hidden states inside the feed-forward network. We quantify the similarity of these projections across different lexical variations of a question using cosine similarity (see \textit{Methods} for a more detailed description of how we quantify this similarity across different lexical variants). The exhaustive list of results across all models and relations can be found in  \textit{Supplementary}, Figure S3.

From Figures \ref{fig:result_neural_capital}(a) and \ref{fig:result_neural_capital}(b), we observe a common trend irrespective of popularity. The similarity of the knowledge retrieved from the model parameters decreases in the early layers, reaching the minima in the middle layers, followed by an increase in the penultimate layers. However, when accounting for popularity in the middle layers, it is evident that the similarity of knowledge retrieved for different lexical variations is inversely related to popularity  This suggests that the higher popularity of a question leads to the LLM becoming more sensitive to linguistic variations in terms of the knowledge retrieved from the parametric memory. Moreover, from Figures \ref{fig:result_neural_capital}(c) and  \ref{fig:result_neural_capital}(d), we observe an increasingly higher probability score of the target token from predicted probability distributions from final layers, as the popularity of the questions increase. This indicates that despite the information anxiety in the knowledge retrieval process, LLMs remain confident in their prediction for high popularity questions. It is this result that is captured by specific high-performance metrics at a macroscopic level.

\subsection*{Similarity of Facts Retrieved among Relations}
\begin{figure}[ht!]
    \centering
    \includegraphics[width=\linewidth]{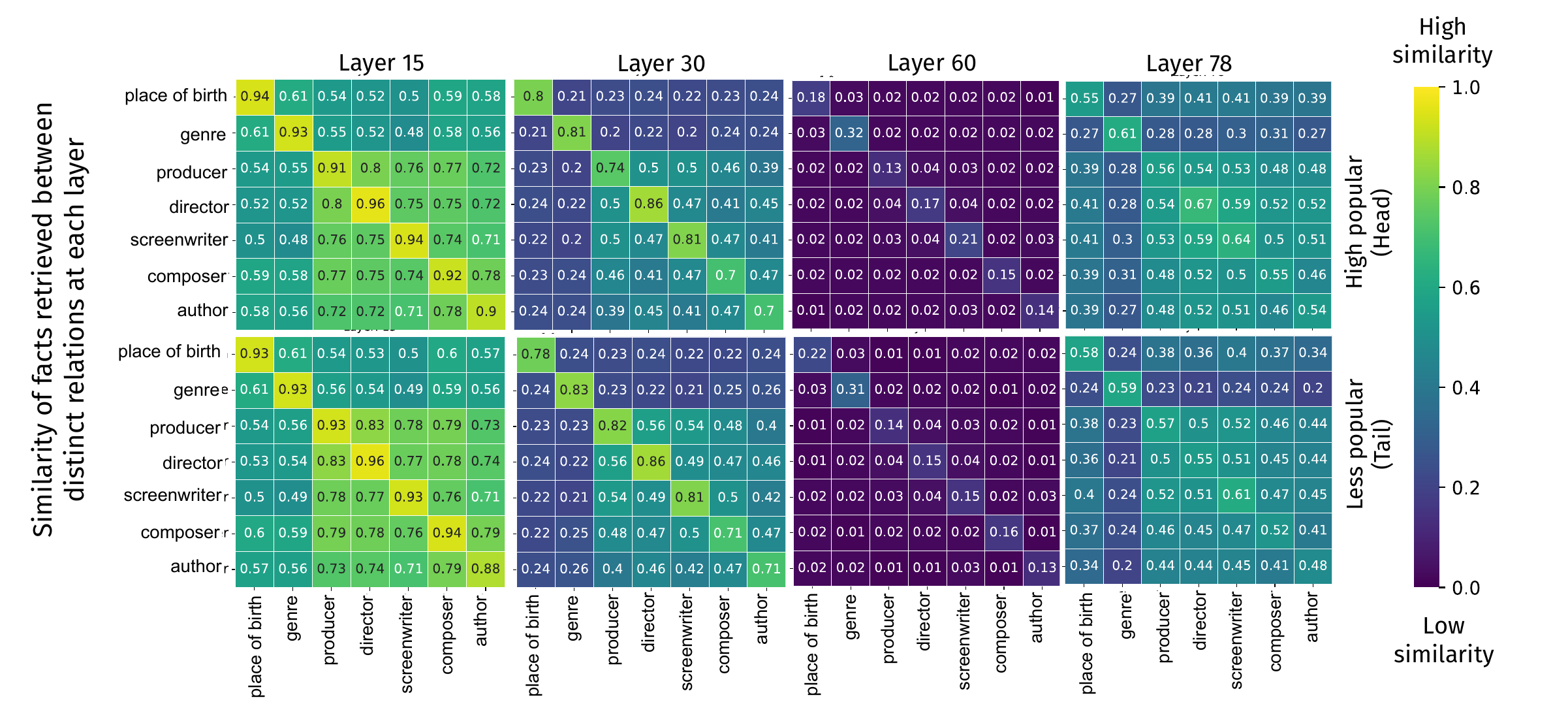}
    \caption{\textbf{Similarity of facts retrieved among relations (in Llama-2-70B)}. For each pair of relations, we show the mean of the similarity of facts retrieved across all questions at each layer. For both the high popular (head) and low popular (tail) questions, the similarity of the knowledge retrieved between any two pairs of relations first decreases across the layers and then increases in the final layers. For any pair of relations, we observe that there is a higher similarity between the facts retrieved across almost all layers for the head batch. We find similar patterns for other LLMs ($Supplementary$, Section 2.4).}
    \label{fig:result_matrixadd_70b}
\end{figure}
Intrigued by the previous results, we perform a granular analysis intersecting popularity, information anxiety, and relation type for the linguistic variants of a query. For a particular layer, we record the mean cosine similarity score between all possible pairs of questions of two relations. Details of the experimental setup are available in \textit{Methods}. 
The 70B variant has the highest volume of pretraining data, containing a richer presentation of both high and low popularity entities in PopQA\cite{touvron2023llama2openfoundation}. Hence, we spotlight our analysis for this RQ with the 70B model.
Figure \ref{fig:result_matrixadd_70b} shows the results in the form of heatmaps, where each entry represents a similarity score between the facts retrieved corresponding to a pair of relations. In the initial layers of processing, highly similar facts are retrieved from the feed-forward network irrespective of the relation pair. As the hidden states are processed through the layers, the similarity of knowledge retrieved becomes more dependent on the relation pair. Overall, the score decreases with increasing layers, up until the final layers, where the similarity of knowledge retrieved becomes high again, irrespective of the relation pair. For a group of relations, $\{$producer, director, screenwriter, composer, author$\}$, the similarity score remains comparatively high within each pair for relations for this group, as compared to a pair with one member relation from this group and the other outside it. The exhaustive list of results across models and relations can be found in \textit{Supplementary}, Figure S4.

Existing studies\cite{10.1145/3626772.3657743,Zhu2023LargeLM} have shown that higher scaled models retrieve diverse knowledge more effectively  We argue that higher-scaled LLMs have certain internal strains, appearing as a limitation in distinguishing popular facts belonging to distinct relations. Mainly focusing on the final layers before the prediction of the first output token, highly popular questions register a higher inter-relation fact similarity score than low popularity questions  We conduct a one-sided Welch's t-test to ascertain the statistical significance of our claim for layer 78 (pre-final layer) as shown in Figure \ref{fig:result_matrixadd_70b}. With $p=0.097$ and 90\% confidence interval, we reject the null hypothesis that the two sets of popularity have comparable similarity scores.  Empirically, the observations suggest that while dealing with highly popular questions, the LLM struggles to effectively separate facts of distinct relations. These results, combined with our observations from previous RQs, signal information anxiety in the hidden states of LLMs when examined at a granular level. 
\section*{Discussion}
Using Wikipedia page view as a proxy for popularity, our study explores the heightened internal strains in the core components of LLMs when handling highly popular queries in an open-domain QA setting. We term this phenomenon \textit{information anxiety}, which we document within the open-source Llama-2 family using PopQA. Our results suggest that the inherent popular queries lead to a different rate of convergence of the predicted probability distribution towards the target/golden distribution. Specifically, the predicted probability distribution associated with highly popular queries converges much more rapidly compared to the case of less popular questions. This faster convergence indicates that the hidden states associated with popular queries undergo reduced computational processing, increasing the risk of producing underdeveloped probability distributions. We also capture this immaturity in Figure \ref{fig:introduction_variety_screenwriter} by prompting semantically similar lexical variants of queries and showing a marked variety in answers recorded for highly popular questions. The notable lack of robustness in LLM's processing of prevalent (w.r.t pretraining corpus) information leaves it vulnerable to adversarial exploits\cite{Shayegani2023SurveyOV}.

To further elucidate the impact of popularity on integral components of knowledge retrieval, we separately analyze the multi-headed self-attention mechanism and the feed-forward networks. Our analysis of attention scores reveals that constraint tokens linked to prevalent subjects receive significantly lower attention scores. As expected, the attention scores also exhibit high sensitivity to lexical perturbations. Both these results point towards the information strain, aka anxiety in the system. A comprehensive mechanistic interpretation\cite{DBLP:journals/tmlr/DuttaSC024, Bereska2024MechanisticIF} of the impact of popularity in reasoning-intensive tasks remains an open avenue for future research. 

Analyzing the feed-forward network projects also reveals an increased variability in the retrieved facts as the popularity of the questions increases. Our analysis reveals a higher similarity score among facts retrieved for highly popular queries spanning different relations. It suggests that large-scale LLMs encounter challenges in effectively distinguishing knowledge across relational boundaries. The persistence of internal strains in both attention and feed-forward components can impact downstream tasks like model editing, which rely on discriminating factual units in an LLM\cite{10.1145/3698590,NEURIPS2023_3927bbdc}. Our observations also call into question the role of training and evaluation strategies of LLMs.

Regarding knowledge retrieval in LLMs, this study furthers the cause in two parallel research areas. Firstly, reducing sensitivity to lexical variations of a knowledge query is highly critical. Recent works\cite{chatterjee-etal-2024-posix, lu-etal-2024-prompts} highlight significant challenges in this area. Secondly, reducing popularity-dependent strains in the internal mechanisms of LLMs needs urgent focus. An enhanced attention-based scoring and a better understanding of factual recall limitations are a part of further research. The observed sensitivity of the knowledge retrieval mechanisms in addressing highly prevalent information also raises concerns about the need for standardized assessment of the pretraining data quality. It is currently challenging to track and standardize the same, given the volume and close-source nature of the raw pertaining data. We conjecture that multiple sources of information about the entities can lead to variations in the retrieved facts. We, therefore, suspect that better scrutiny of the pretraining corpus might also reduce information anxiety. 

It is imperative to point out that entity and popularity formatted datasets such as PopQA are hard to curate. Developing better popularity proxies for pertaining entities is rather non-trivial. On the one hand, we can take advantage of the hyperlinked and connected structure of Wiki articles \cite{yasunaga-etal-2022-linkbert}. On the other hand, Wikipedia is only one of the many sources of information that are edited using a specific timestamp. Via the use case of information anxiety, our study motivates researchers to explore alternative proxies for the popularity/frequency of entities in the pretraining corpus.
\section*{Methods}

\subsection*{Background}
The transformer decoder\cite{10.5555/3295222.3295349} is stacked with $L$ identical blocks. Each transformer decoder block contains a multi-headed self-attention (MHSA) module and a feed-forward network (FFN) module.
When processing input $x$, the action of a FFN is defined as, 
\begin{equation}
    \text{FFN}(x) = (\text{SiLU}(W_{gate}(x)) \cdot W_{up}(x))W_{down},
\end{equation}
\noindent where $x \in \mathbb{R}^{l \times d_{model}}$ is the hidden state stream fed into the FFN, and $W_{gate}$, $W_{up} \in \mathbb{R}^{d_{model} \times d_{inter}}$ and $W_{down} \in \mathbb{R}^{d_{inter} \times d_{model}}$. For an arbitrary scalar $y$, SiLU(y)\cite{DBLP:journals/corr/ElfwingUD17} is defined as,
\begin{equation}
    \text{SiLU}(y) = y \cdot \frac{1}{1 + e^{-y}}.
\end{equation}
The MHSA module in a certain layer is defined as,
\begin{equation}
    \text{MHSA}(x) = \text{Softmax}(\frac{W_{Q}(x)W_{K}^{T}(x)}{\sqrt{d}})W_{V}(x),
\end{equation}
where $W_{Q}$, $W_{K}$ and $W_{V} \in \mathbb{R}^{d_{model} \times d_{mid}}$. The values $d_{model}$, $d_{inter}$ and $d_{mid}$ are model-specific dimensional hyperparameters. For an arbitrary vector $\vec{y} = (y_1, y_2, ..., y_n)$, Softmax($\vec{y}$) is defined as,
\begin{equation}
    \text{Softmax}(\vec{y}) = (\frac{e^{y_1}}{\sum_j e^{y_j}}, \frac{e^{y_2}}{\sum_j e^{y_j}}, ..., \frac{e^{y_n}}{\sum_j e^{y_j}})^T.
\end{equation}
In the subsequent sections, we detail the methodology of each study, focusing on its specific scope in addressing \textit{information anxiety}. Our methods are backed by recent studies\cite{lv2024interpretingkeymechanismsfactual,dai-etal-2022-knowledge} delving into critical interpretations of factual recall mechanisms in transformer-based LLMs, where the attention heads extract key tokens from a query and the corresponding FFNs direct the hidden states towards the correct answer. In this study, we look closely at both of these steps in the overall factual processing by reporting the effect of popularity on LLMs' capability to attend to key tokens in a factual query and consequently retrieve facts from the parameters of the FFN. Furthermore, this study offers an additional perspective on how sensitive these factual recall mechanisms are to variations in the lexical structure of queries across different popularity scales.

\subsection*{Dataset} 
We use PopQA\cite{mallen-etal-2023-trust}, an entity-centric open-domain QA dataset consisting of 14k questions, to cover factual information pertaining to both highly and less popular subjects that might otherwise have been overlooked in other widely used QA datasets\cite{kwiatkowski-etal-2019-natural,joshi-etal-2017-triviaqa}. PopQA has QA pairs grounded with fine-grained Wikidata entity ID\cite{10.1145/2629489}, Wikipedia page views, and relationship-type information. Each query in PopQA is annotated with the popularity of the subject ($p_{subj}$) and the popularity of the object ($p_{obj}$), which is based on the corresponding entity's Wikipedia page view cumulative numbers. Knowledge triplets from Wikidata, with diverse levels of popularity, are then provided as natural language questions anchored to the original entities and relationship types. To sort the questions based on popularity, we use the mean $(p_{obj} + p_{subj}) / 2$ as the comparator. In this text, the popularity of a question refers to the average of the subject's ($p_{subj}$) and the target object's ($p_{obj}$) popularity.

\subsection*{Lexical Variations}
\label{sec:methods-paraphrase-templates}
For each relation in PopQA, we curate ten (including the original query) different paraphrases of the original query. Each of these paraphrases represents only a lexical variant of the original query while keeping the underlying semantic sense of the query intact. We ensure that all of the different paraphrases point to the same answer to account for only the effect of lexical variations on the reasoning and fact retrieval structures of the models analyzed. For example, if the original question was `\textit{What is the place of birth for} <SUBJECT>?', we include the paraphrase `\textit{Where was} <SUBJECT> \textit{born}?'. All the paraphrases corresponding to each relation in PopQA can be found in \textit{Supplementary}, Table S1.

\subsection*{Experimental Setup}
We conduct our experiments on the following open-sourced models of Llama-2\cite{touvron2023llama2openfoundation} family: Llama-2-7B, Llama-2-7B-chat, Llama-2-13B, Llama-2-13B-chat and Llama-2-70B. We prompt LLMs in an in-context setting\cite{Radford2019LanguageMA}, randomly sampling 16 QA pairs belonging to the same relation as the target query. We use these sampled QA pairs as demonstrations in the prompt structure, with `\textit{Answer the following question in one word or phrase.}' as the task descriptor. The number of demonstrations is arbitrarily selected from observing the performance of models that are not instruction fine-tuned. The exact prompt structure is presented in \textit{Supplementary}, Section S1.2.

\subsection*{Convergence of the Predicted Probability Distribution}
\label{sec:methods_convergence_of_the_probability_distribution}
Let $L$ be the total number of identical decoder blocks in the transformer model under study. We define $h_{l}$ to be the hidden representations after processing through the $l^{th}$ decoder block,
\begin{equation}
    h_{l} = \text{FFN}(\text{Attention}(h_{l-1})), \hspace{0.2cm} 1 \leq l \leq L
\end{equation}
where $h_{0}$ is the initial hidden representation. The Llama-2 family has an internal embedding lookup matrix $E \in \mathbb{R}^{|V| \times d_{model}}$, where $|V|$ represents the vocabulary size. This lookup matrix is used to obtain the initial hidden representation $h_{0}$ from the input token IDs. For layer $l$, the hidden representation $h_{l}$ can be used similarly to obtain a probability distribution over the vocabulary space, denoted by $p^{l}$. The final output token $o$ is sampled from the probability distribution $p^{L}$: $o \sim p^{L}$.
For each relation $r$, we divide the subset for that relation into $N = 5$ batches : $\{B_i\}_{i = 1}^{N}$ with each batch containing 50 questions (i.e., $|B_{i}|$ = 50). We order the batches $\{B_i\}$ such that $B_1$ has the set of highest popularity questions of that relation while $B_{N}$ has the set of least popular questions. We label $B_1$ as the \texttt{Head} batch, $B_{3}$ as \texttt{Torso} and the batch $B_{5}$ as the \texttt{Tail} set of questions.\par
For each question $Q_i$, we prompt each of the $p = 10$ lexical variations, denoted by $P_{i}^{j}: 1 \leq j \leq p$. For each $P_{i}^{j}$, we intercept the hidden states $h_{l}: 0 \leq l \leq L$, where $L$ is the total number of layers of the model under study. We obtain the predicted probability distribution for each layer, $p^{l}$, by taking the product $p^{l} = E^{T}h_{l}$, where $E$ is the embedding lookup matrix. For each question $Q_i$, we also have a target answer $A_{i}$, and we use the tokenizer to obtain the first sub-word token of $A_i$, say $a_i$. We create the golden (or target) probability distribution $\Tilde{p}$ by assigning a full mass of 1 to $a_i$ in the vocabulary space. We then calculate the KL-Divergence between $p^{l}$ and $\Tilde{p} \hspace{0.2cm} \forall \hspace{0.2cm} 0 \leq l \leq L$ for the $j^{th}$ paraphrase lexical variation as follows:
\begin{equation}
    D_{l}^{j}(\Tilde{p} || p^{l}) = \sum_{k} (\Tilde{p}_k log \frac{\Tilde{p}_k}{p^{l}_k}).
\end{equation}
Note $D_{l}^{j}$ corresponds to the $j^{th}$ lexical variation $P_{i}^{j}$ for the $i^{th}$ question $Q_{i}$. We calculate the mean ($_{prob}M_i^l$) and standard deviation ($_{prob}V_i^l$) across all $p = 10$ lexical variants, i.e., for question $Q_i$ at $l^{th}$ layer,
\begin{equation}
    _{prob}M_i^l = \frac{\sum_{j=1}^{p}D_{l}^{j}}{p}; \hspace{0.3cm} _{prob}V_i^l = \sqrt{\frac{\sum_{j=1}^{p}(D_{l}^{j} - _{prob}M_i^l)^{2}}{p}}.
\end{equation}
\noindent
For each defined batch of questions, \texttt{Head}, \texttt{Torso} and \texttt{Tail}, we plot the average of $_{prob}M_i^l$ and $_{prob}V_i^l$ to get a batch representation of statistics defined above. For instance, in the case of batch \texttt{Head}:
\begin{equation}
    _{prob}M^l = \frac{\sum_{i} {_{prob}M_i^l}}{|\texttt{Head}|}, \hspace{0.3cm} _{prob}V^l = \frac{\sum_{i} {_{prob}V_i^l}}{|\texttt{Head}|}.
\end{equation}
We plot $_{prob}M^l$ and $_{prob}V^l$ versus layer $l$ in Figure \ref{fig:result_distance-hidden_capital_genre} for the `Capital' and `Genre' relations.

\subsection*{Effect of Popularity on LLM Reasoning Behaviour}
\label{sec:effect_of_popularity_on_llm_reasoning_structure}

Building upon previous studies\cite{hou2023mechanisticinterpretationmultistepreasoning,stolfo2023mechanisticinterpretationarithmeticreasoning,friedman2024interpretabilityillusionsgeneralizationsimplified} that link self-attention structures to reasoning, we explore the effect of popularity on the internal reasoning mechanisms by focusing on the MHSA layer. Mainly, we account for the impact of the popularity of the question on the ability of LLMs to attend to different parts of the query. To scope in on the significant portions of the query, we label some tokens as constraint tokens for each lexical variant of all questions in PopQA. For example, the constraint tokens for the question `What is the \textit{capital} of \textit{Ireland}?', the constraint tokens are \textit{italicized}. We divide each relation-wise subset of PopQA into $N = 2$ batches of 50 questions each. We denote by \texttt{Head} as the batch of most popular questions and \texttt{Tail} as the batch of least popular questions. \par
Let us consider a question $Q_i$ with each of its $p = 10$ paraphrases $P_i^j, 1 \leq j \leq p$. For each paraphrase $P_i^j$, we define set $C(P_i^j)$ to be the set of constraint tokens. At a layer $l$, for each constraint token $c \in C(P_i^j)$, we take the average of attention scores across all attention heads. We record the maximum of these average attention scores across all constraint tokens for each paraphrase template. At a layer $l$, for each constraint token $c \in C(P_i^j)$, we record the attention
\begin{equation}
    \text{Attn-score}_l(P_i^j) = \max_{c \in C(P_i^j)}\frac{\sum_h \text{Attn}_l^h(c)}{|H|},
\end{equation}
where $\text{Attn}_l^h(\cdot)$ represents attention score due to the attention head $h$ in layer $l$, and $|H|$ denotes the total number of attention heads. At a layer $l$, for each question $Q_{i}$, we record the average ($_{attn}M_i^l$) of $\text{Attn-score}_l(\cdot)$ and standard deviation ($_{attn}V_i^l$) across all paraphrases,
\begin{equation}
    _{attn}M_i^l = \frac{\sum_{j=1}^{p}\text{Attn-score}_l(P_i^j)}{p}, \hspace{0.2cm} _{attn}V_i^l = \sqrt{\frac{\sum_{j=1}^{p}(\text{Attn-score}_l(P_i^j) - {_{attn}M_i^l})^{2}}{p}}.
\end{equation}
For a batch of questions, say \texttt{Head}, we take a batch representation of the above-defined statistics by taking the average across all questions in that batch,
\begin{equation}
    _{attn}M^l = \frac{\sum_{i} {_{attn}{M_i^l}}}{|\texttt{Head}|}, \hspace{0.3cm} _{attn}V^l = \frac{\sum_{i} {_{attn}{V_i^l}}}{|\texttt{Head}|}. 
\end{equation}
We plot $_{attn}M^l$ and $_{attn}V^l$ versus layer $l$ in Figure \ref{fig:result_attention_captial_occupation} for `Occupation' and `Capital' relation and models, Llama-2-7B and Llama-2-7B-chat, in the form of solid line and error bar, respectively.

\subsection*{Effect of Popularity in LLM Fact Retrieval Behaviour}
Studies \cite{geva2021transformerfeedforwardlayerskeyvalue,dai-etal-2022-knowledge,meng2023locatingeditingfactualassociations} have shown that transformer feed-forward networks are stores of knowledge and are fundamental for fact retrieval. We seek to find if an increase in the popularity of a question causes any difference in the facts retrieved from FFNs when prompted with lexical variations of questions. We divide each relation-wise subset of PopQA into $N = 5$ batches of 50 questions each.\par
Let us consider a question $Q_i$ with each of its $p = 10$ paraphrase templates $P_i^j, 1 \leq j \leq p$. For each paraphrase $P_i^j$ at each layer $l: 1 \leq l \leq L$, where $L$ is the number of layers in the model considered, we intercept the upward projections in each FFN structure (denoted by $\text{Up-Proj}_j^l$):
\begin{equation}
    \text{Up-Proj}_j^l = W_{up}(h_{l-1}).
\end{equation}
For each of the $p = 10$ paraphrase templates, we take the average of the pair-wise cosine similarity scores between the intercepted upward projections and label it as $\text{Sim}_i^l$ to denote the similarity between the upward projections for $p = 10$ paraphrase templates for the $i^{th}$ question at the $l^{th}$ layer.
\begin{equation}
    \text{Sim}_i^l = \frac{\sum_j \sum_k \text{cosine-similarity}(\text{Up-Proj}_j^l, \text{Up-Proj}_k^l)}{p^2}.
\end{equation}
We obtain the batch representation (say for a batch $B$) of the statistics defined above by taking an average of $\text{Sim}_i^l$ for each question $Q_i$.
\begin{equation}
    \text{Sim}_l = \frac{\sum_i \text{Sim}_i^l}{|\text{B}|},
\end{equation}
where $|B|$ ($= 50$) is the number of questions in each batch. 
To showcase that LLMs are increasingly confident of their predictions at higher levels of popularity, for each question $Q_i$, we store the average of probability scores across all paraphrase templates for the correct target token from the predicted probability distribution at each layer. Let us denote this quantity by $p_i^l$. We obtain the batch representation of this probability score by taking an average of $p_i^l$ for each question $Q_i$,
\begin{equation}
    p_i = \frac{\sum_i p_i^l}{|{B}|}.
\end{equation}
We plot $\text{Sim}_l$ and $p_l$ versus each layer $l$ in Figure \ref{fig:result_neural_capital} for the `Capital' relation and two models, Llama-2-7B and Llama-2-13B-chat.

\subsection*{Similarity of Knowledge Retrieved among Relations}

We sort the entire PopQA dataset in the decreasing order of the popularity of questions. We divide the entire dataset into $N = 5$ batches of $2,000$ questions each such that the batches are equispaced in terms of the average popularity of the questions. Since the range of popularity of questions for each relation $r$ in PopQA is different, we find that only relations \{`place of birth,' `genre,' `producer,' `director,' `screenwriter,' `composer,' `author'\} occur in each of the batch with at least 50 questions per relation. Due to this compositional limitation of PopQA, we limit our analysis on this collection of relations. Let us denote this collection of relations as $\{\Bar{R}\}$. Let $B^{\Bar{R}}_{i}$ denote the $i^{th}$ batch of the relation $R$ (|$B^{\Bar{R}}_{i}$| = 50). The batches are indexed in such a way that $i \leq j$ implies that $B^{\Bar{R}}_{i}$ contains a set of more popular questions than $B^{\Bar{R}}_{j}$. Note that this way of creating batches ensures that for each batch level, the questions are of almost equal popularity irrespective of the relation.
For each batch, we seek to find if there is a similarity in the knowledge retrieved between any two arbitrary relations.\par
For two relations $\Bar{R}_x$ and $\Bar{R}_y$, we sample one question from $B^{\Bar{R}_x}_{i}$ and a question from $B^{\Bar{R}_y}_{i}$ and calculate the similarity of the upward projections at each layer of processing. Consequently, at each layer, we average out the similarity across all possible choices of questions ($50\choose2$) from $B^{\Bar{R}_x}_{i}$ and $B^{\Bar{R}_y}_{i}$. So, for $i^{th}$ level of batches, for relations $\Bar{R}_x$ and $\Bar{R}_y$ at each layer $l$, we denote the similarity of knowledge retrieved by $\text{Sim-Know}(\Bar{R}_x, \Bar{R}_y)_{l}^{i}$ and we define it by:
\begin{equation}
    \text{Sim-Know}(\Bar{R}_x, \Bar{R}_y)_{l}^{i} = \sum_{q_x \in B^{\Bar{R}_x}_{i}} \sum_{q_y \in B^{\Bar{R}_y}_{i}} \text{cosine-similarity}(\text{Up-Proj}_{l}(q_x), \text{Up-Proj}_{l}(q_y)) / {50 \choose 2},
\end{equation}
where $\text{Up-Proj}_{l}(\cdot)$ represents the upward projections of the $l^{th}$ layer for the corresponding question.
\bibliography{sample}

\section*{Author Contributions} 
P.B. and T.C. contributed to conceptualizing and leading the effort of writing the initial draft of the manuscript. P.B., T.C. and S.M. prepared the final version of the manuscript. T.C. coordinated the entire project.

\section*{Acknowledgment}
T.C. acknowledges the support of IBM AI Horizons
Network (AIHN) grant and Rajiv Khemani Young Faculty Chair in AI fellowship. S.M. is financially supported by Google India PhD Fellowship. All the authors acknowledge Dr Subhabrata Dutta for his insights in formulating the problem statement.

\section*{Competing Interests}
The authors declare no competing interests.

\section*{Additional Information}

\noindent{Materials \& Correspondence} should be emailed to Tanmoy Chakraborty (\url{tanchak@iitd.ac.in}).
\newpage
\section*{Supplementary Information}
\section*{S1 Methods}
\subsection*{S1.1 Lexical Variants}
For each of the relations in \texttt{PopQA}, we curate 10 lexical variants of the original query while preserving the semantic sense of the query. We highlight the constraint tokens in blue, for each of the lexical variants.
\begin{longtable}{p{0.2\linewidth}|@{\hspace{0.8cm}}p{0.7\linewidth}}
\textbf{\texttt{capital}} & \begin{tabular}{l}
    What is the \format{capital} of \format{<SUBJECT>}?\\
    What is the \format{primary city} in \format{<SUBJECT>}?\\
    Which \format{urban center} holds a \format{central role} in \format{<SUBJECT>}?\\
    Which city \format{houses the main government} of \format{<SUBJECT>}?\\
    What city is the \format{main administrative center} of \format{<SUBJECT>}?\\
    What \format{urban area} holds the \format{capital status} of \format{<SUBJECT>}?\\
    Which city is the \format{political focal point} of \format{<SUBJECT>}?\\
    Which city is the \format{primary seat of authority} for \format{<SUBJECT>}?\\
    What is the \format{principal city} of \format{<SUBJECT>}?\\
    What is the \format{chief city} of \format{<SUBJECT>}?\\
    \end{tabular}\\\\
\textbf{\texttt{genre}} & \begin{tabular}{l}
    What \format{category} does \format{<SUBJECT>} belong to?\\
    What \format{kind of classification} does \format{<SUBJECT>} fall under?\\
    What \format{type} does \format{<SUBJECT>} fall under?\\
    What \format{group} does \format{<SUBJECT>} fall under?\\
    What \format{kind} does \format{<SUBJECT>} fall under?\\
    What \format{class} does \format{<SUBJECT>} fall under?\\
    What \format{style} can \format{<SUBJECT>} be categorized into?\\
    Which \format{division} does \format{<SUBJECT>} come under?\\
    In what \format{bracket} does \format{<SUBJECT>} reside?\\
    What \format{domain} does  \format{<SUBJECT>} fit within?\\
\end{tabular}\\\\
\textbf{\texttt{occupation}} & \begin{tabular}{l}
    What does \format{<SUBJECT>} do for a \format{living}?\\
    What is \format{<SUBJECT>}'s \format{occupation}?\\
    What is \format{<SUBJECT>}'s \format{profession}?\\
    What is \format{<SUBJECT>}'s \format{line of work}?\\
    What \format{job role} is \format{<SUBJECT>} engaged in?\\
    What \format{career path} has \format{<SUBJECT>} chosen?\\
    What \format{field of employment} is \format{<SUBJECT>} part of?\\
    What is \format{<SUBJECT>}'s source of \format{professional engagement}?\\
    What is \format{<SUBJECT>}'s \format{vocational pursuit}?\\
    What \format{field of work} is \format{<SUBJECT>} engaged in?\\
\end{tabular}\\\\
\textbf{\texttt{religion}} & \begin{tabular}{l}
    What is the \format{religion} of \format{<SUBJECT>}?\\
    What \format{faith} does \format{<SUBJECT>} follow?\\
    What is the \format{religious affiliation} of \format{<SUBJECT>}?\\
    Which \format{religious doctrine} does \format{<SUBJECT>} subscribe to?\\
    What \format{theology} is \format{<SUBJECT>} aligned with?\\
    What is the \format{religious identity} of \format{<SUBJECT>}?\\
    What \format{belief system} does \format{<SUBJECT>} ascribe to?\\
    What \format{religious school of thought} does \format{<SUBJECT>} follow?\\
    What is \format{<SUBJECT>}'s \format{spirituality}?\\
    What is the \format{religious allegiance} of \format{<SUBJECT>}?\\
\end{tabular}\\\\
\textbf{\texttt{screenwriter}} & \begin{tabular}{l}
        Who was the \format{screenwriter} for \format{<SUBJECT>}?\\
        Who \format{wrote the screenplay} for \format{<SUBJECT>}?\\
        Who was the \format{scribe of the screenplay} in \format{<SUBJECT>}?\\
        Who took on the \format{role of crafting} \format{<SUBJECT>}'s \format{script}?\\
        Who was responsible for \format{composing the screenplay} of \format{<SUBJECT>}?\\
        Who was the \format{wordsmith behind the screenplay} of \format{<SUBJECT>}?\\
        Who \format{penned the script} for \format{<SUBJECT>}?\\
        Who \format{authored the screenplay} of \format{<SUBJECT>}?\\
        Who was the primary influencer in shaping \format{<SUBJECT>}'s \format{screenplay composition}?\\
        Who \format{scripted} \format{<SUBJECT>}'s story?\\
\end{tabular}\\\\
\textbf{\texttt{sport}} & \begin{tabular}{l}
        What \format{sport} does \format{<SUBJECT>} play?\\
        What \format{athletic activity} does \format{<SUBJECT>} participate in?\\
        What \format{game} does \format{<SUBJECT>} compete in?\\
        What \format{field of sports} is \format{<SUBJECT>} involved with?\\
        Which \format{athletic endeavour} does \format{<SUBJECT>} pursue?\\
        What \format{physical activity} does \format{<SUBJECT>} perform as a sport?\\
        What \format{recreational sport} does \format{<SUBJECT>} enjoy?\\
        What \format{sport} does \format{<SUBJECT>} enjoy playing?\\
        What \format{professional sport} does \format{<SUBJECT>} play?\\
        What \format{game} does \format{<SUBJECT>} actively participate in?\\
\end{tabular}\\\\
\caption*{\textbf{Table S1}. For each relation $r$ of \texttt{PopQA}, we curate 10 different lexical variations of the original query while preserving the semantic sense of the original query. Moreover, for focussing on attention scores over a given set of tokens, we \textit{italicize} constraint tokens for each of the lexical variants of a query.}
\end{longtable}

\subsection*{S1.2 Prompt Structure}
Below is the prompt structure we utilize for retrieving responses from all models under study. We sample 16 demonstrator question-answer pairs randomly for each target query ensuring they belong to the same relation.\\\\
\medskip
\medskip
\noindent
\texttt{
Answer the following questions in one word or phrase:\\
Q: <QUESTION 1>\\
A: <ANSWER 1>\\
...\\
Q: <QUESTION 16>\\
A: <ANSWER 16>\\
Q: <\textit{TARGET QUERY}>\\
A: <\textit{TO BE FILLED BY MODEL}>\\
}
\medskip



\section*{S2 Results}
\subsection*{S2.1 Convergence of the Predicted Probability Distribution}
We provide results across all models and relations for checking the convergence of the predicted probability distribution to the golden probability distribution. The results show that the predicted probability distribution approaches the golden probability distribution in the middle layers ($\sim 10-15^{th}$ layer for Llama-2-7B (32 layers), $\sim 15-20^{th}$ layer for Llama-2-13B (40 layers) \& $\sim 30-40^{th}$ layer for Llama-2-70B (80 layers)). 
\begin{longtable}{c} 
\includegraphics[width = 0.95\linewidth]{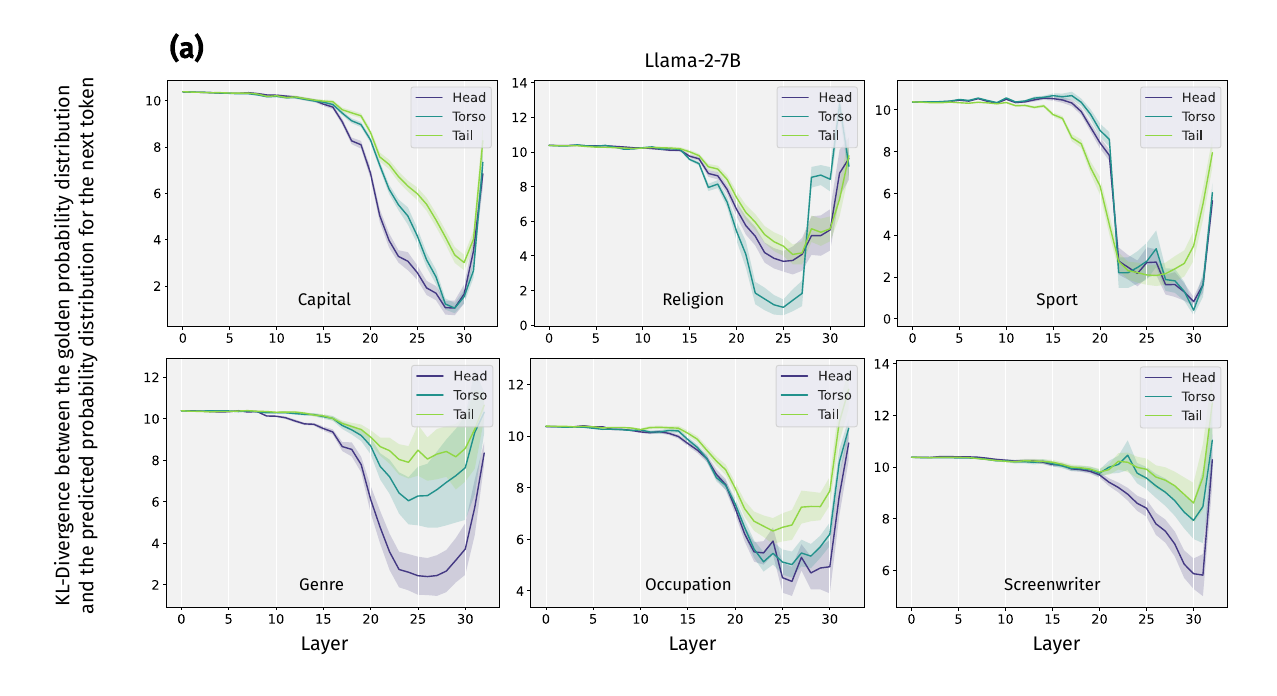} \\
\includegraphics[width = 0.95\linewidth]{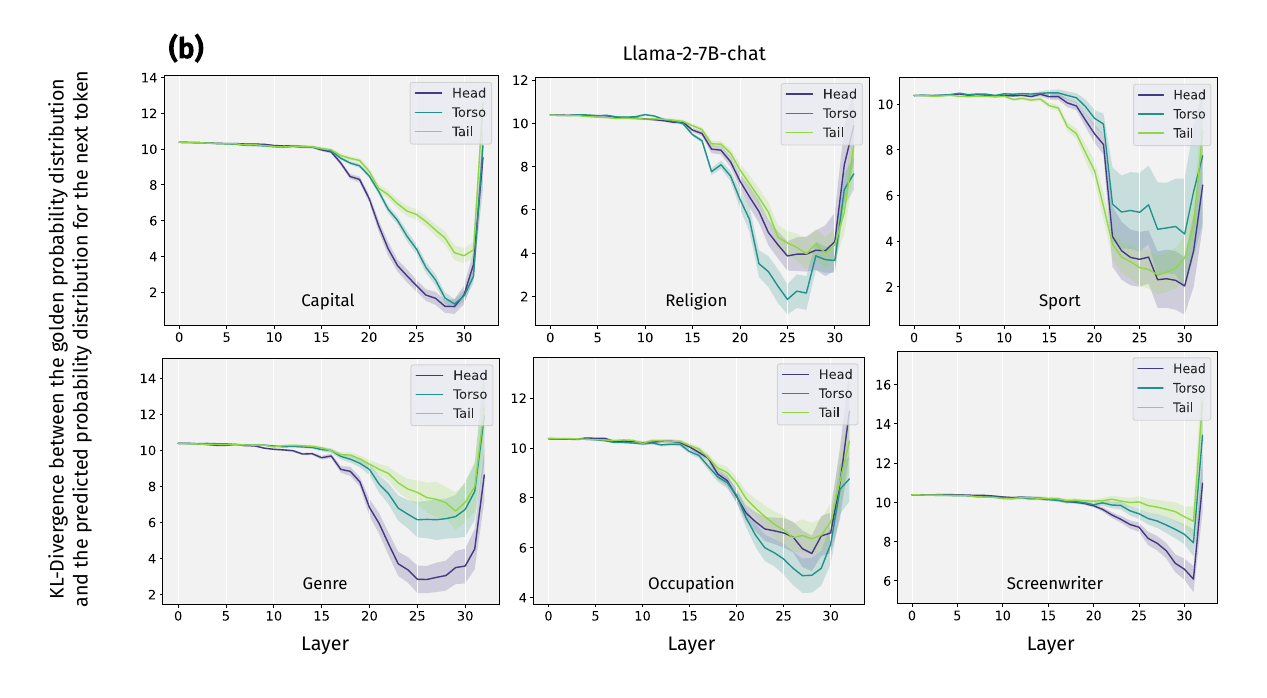} \\
\includegraphics[width = 0.95\linewidth]{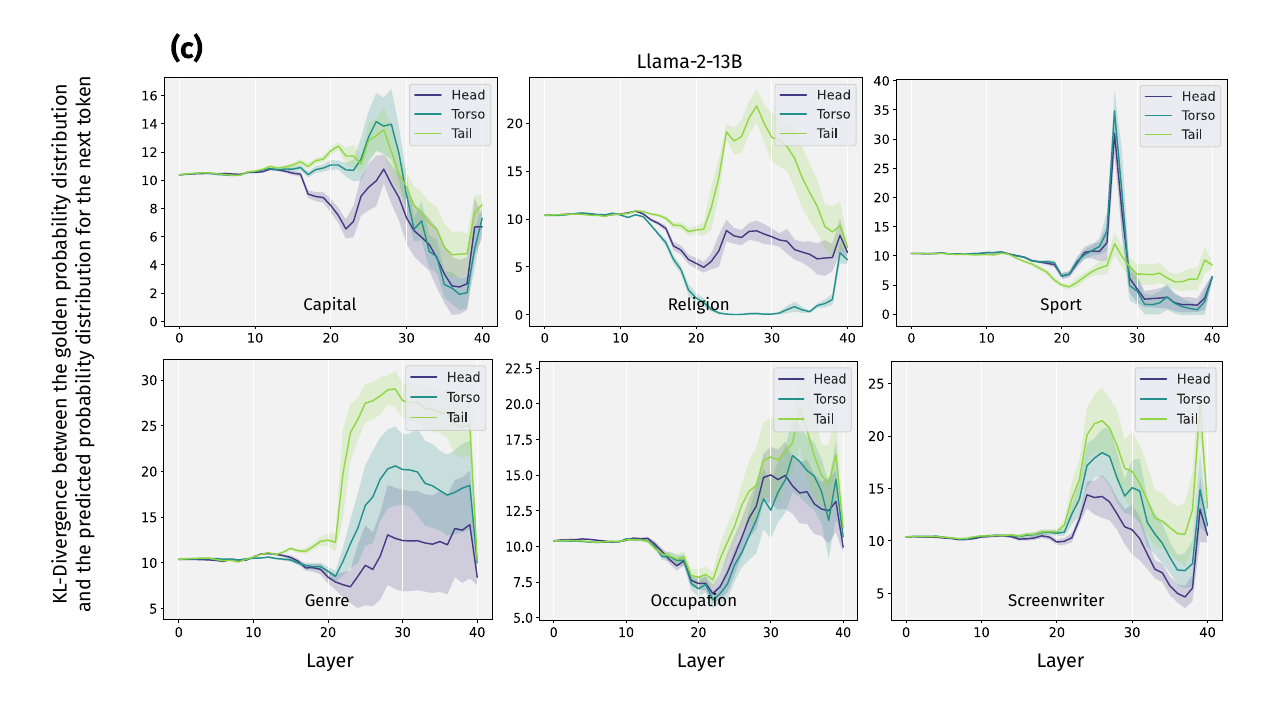} \\
\includegraphics[width = 0.95\linewidth]{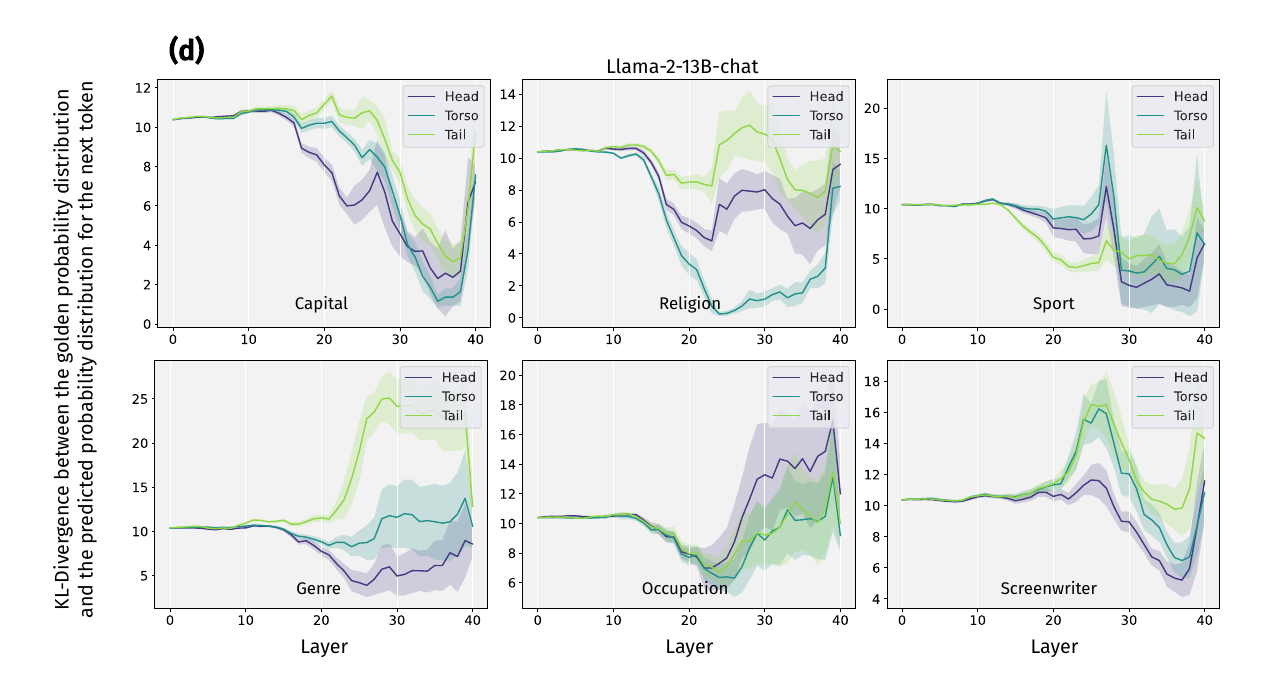} \\
\includegraphics[width = 0.95\linewidth]{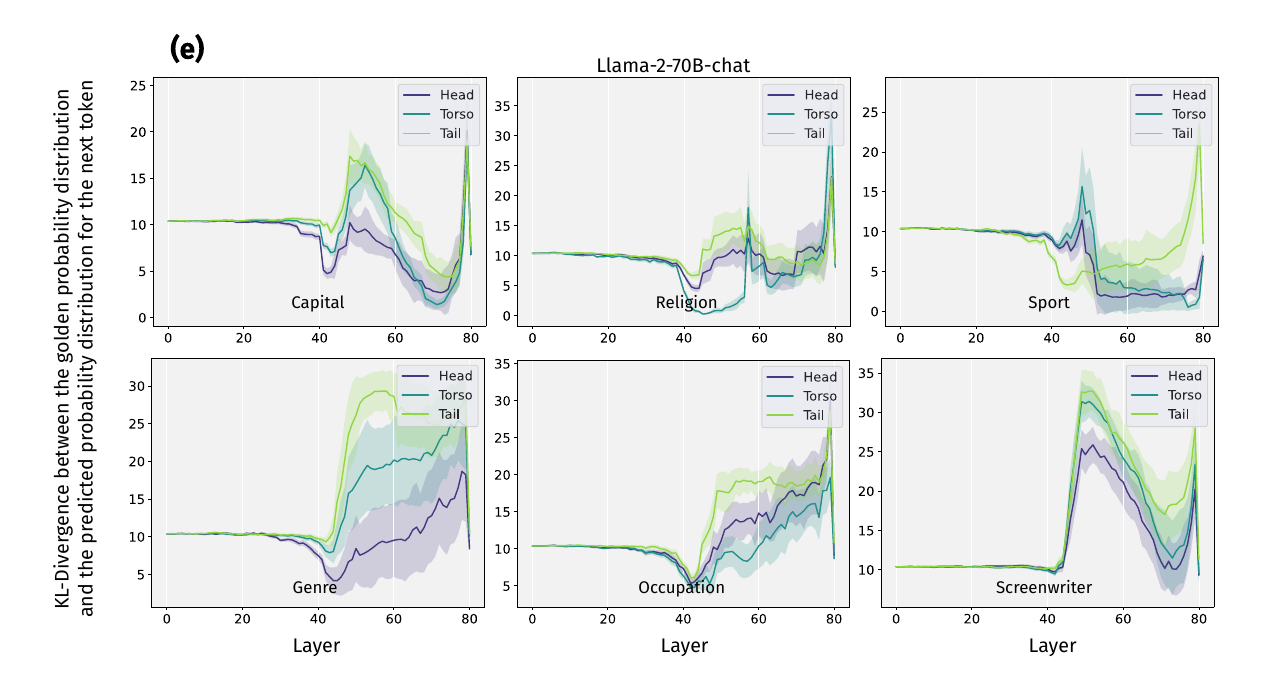} \\
\caption*{\textbf{Figure S1.} \textbf{Convergence of the predicted probability distribution}. The figure shows KL-Divergence between the next
token predicted probability distribution and the golden probability distribution for questions of capital and genre relation
in PopQA. For each batch of questions, the solid line denotes the average of the distances between the predicted and the golden
distributions across all questions in a batch ({\tt Head}, {\tt Torso} or {\tt Tail}). We also account for the effect of lexical variations on
the next token probability distribution. The shaded regions denote the average impact of such variations (captured using
deviations in the KL-Divergence distances) across all questions in a batch. We observe the predicted probability distribution
associated with highly popular questions approaches the golden probability distribution earlier as compared to that associated
with comparatively less popular questions.}
\end{longtable}

\subsection*{S2.2 Effect of Popularity in LLM Reasoning Behaviour}
We provide results across all models and relations for assessing the effect of popularity on the internal reasoning mechanisms of LLMs.
\begin{longtable}{c} 
\includegraphics[width = 0.95\linewidth]{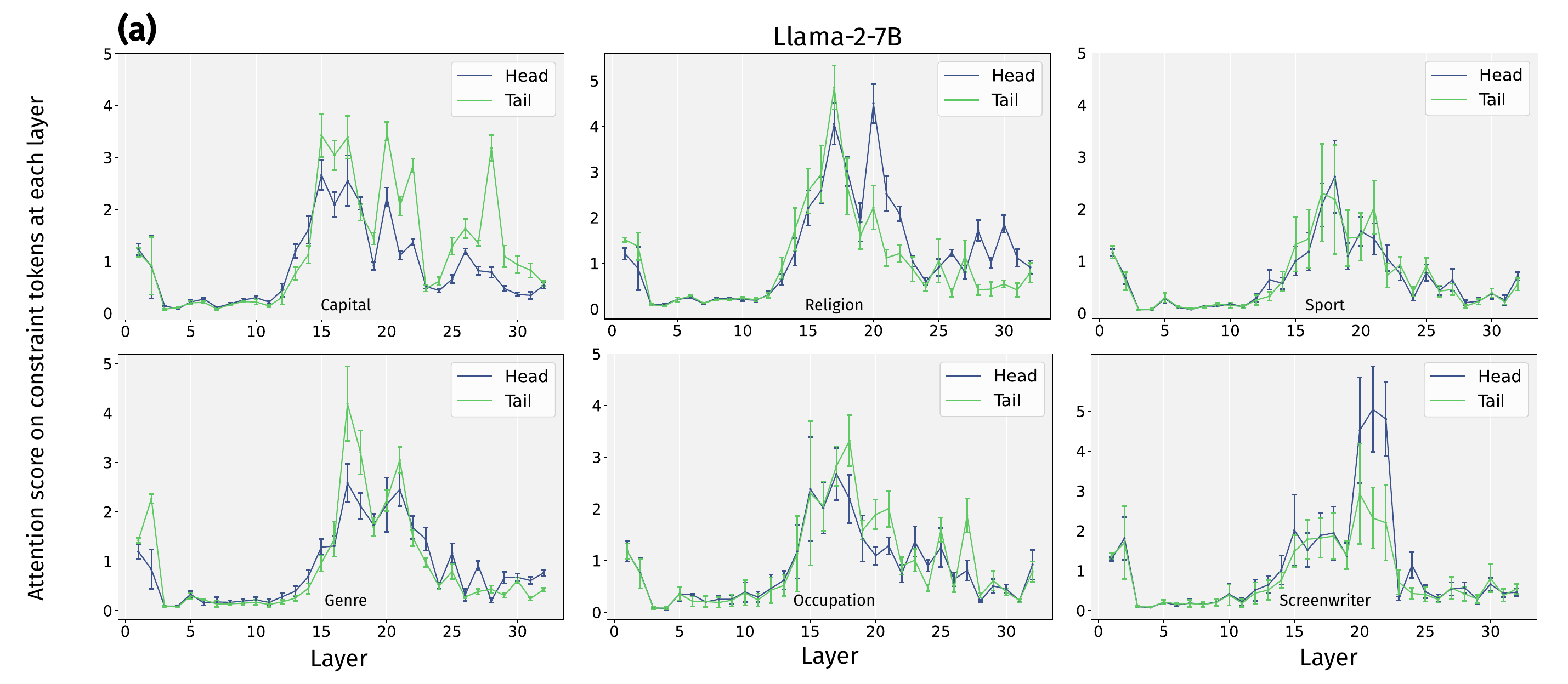} \\
\includegraphics[width = 0.95\linewidth]{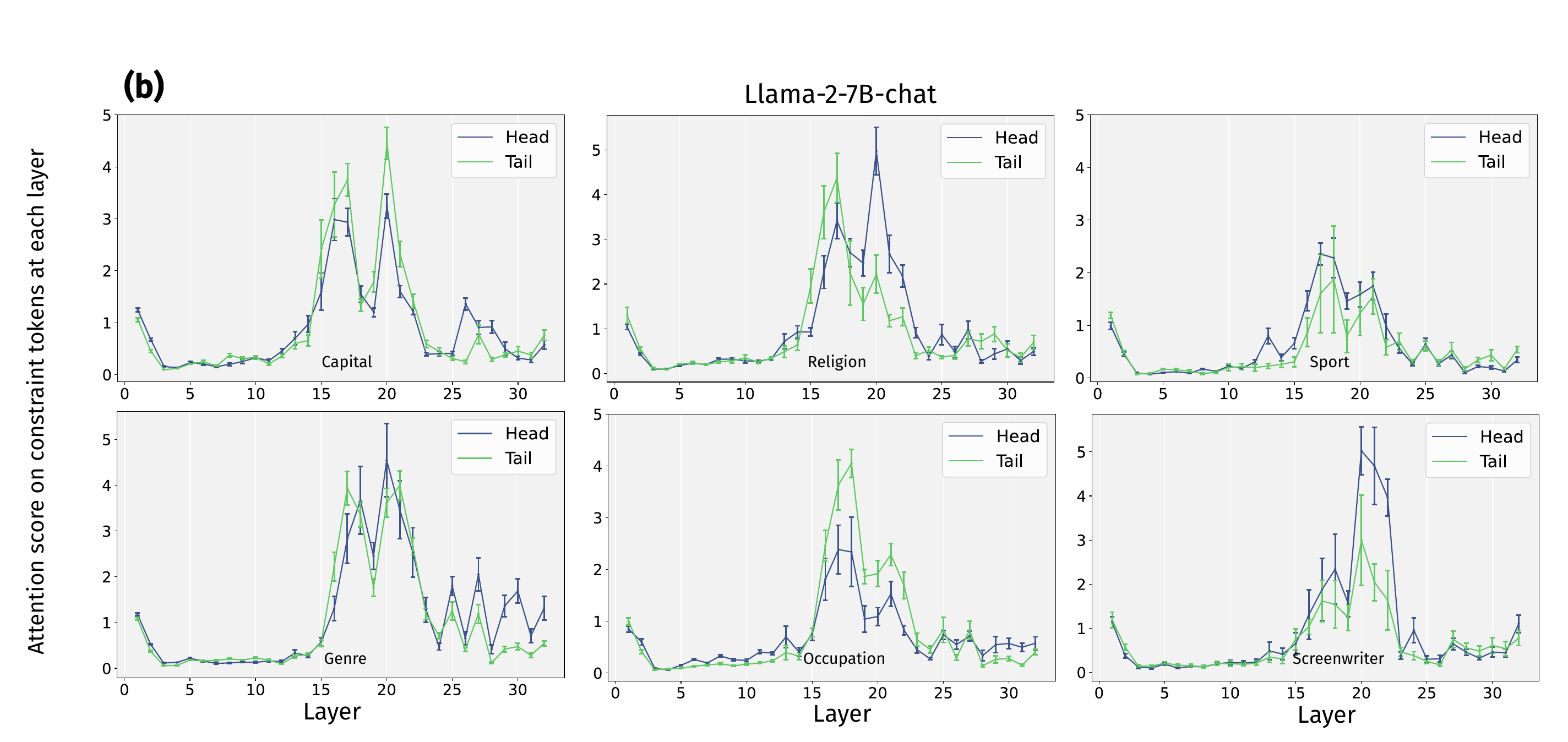} \\
\includegraphics[width = 0.95\linewidth]{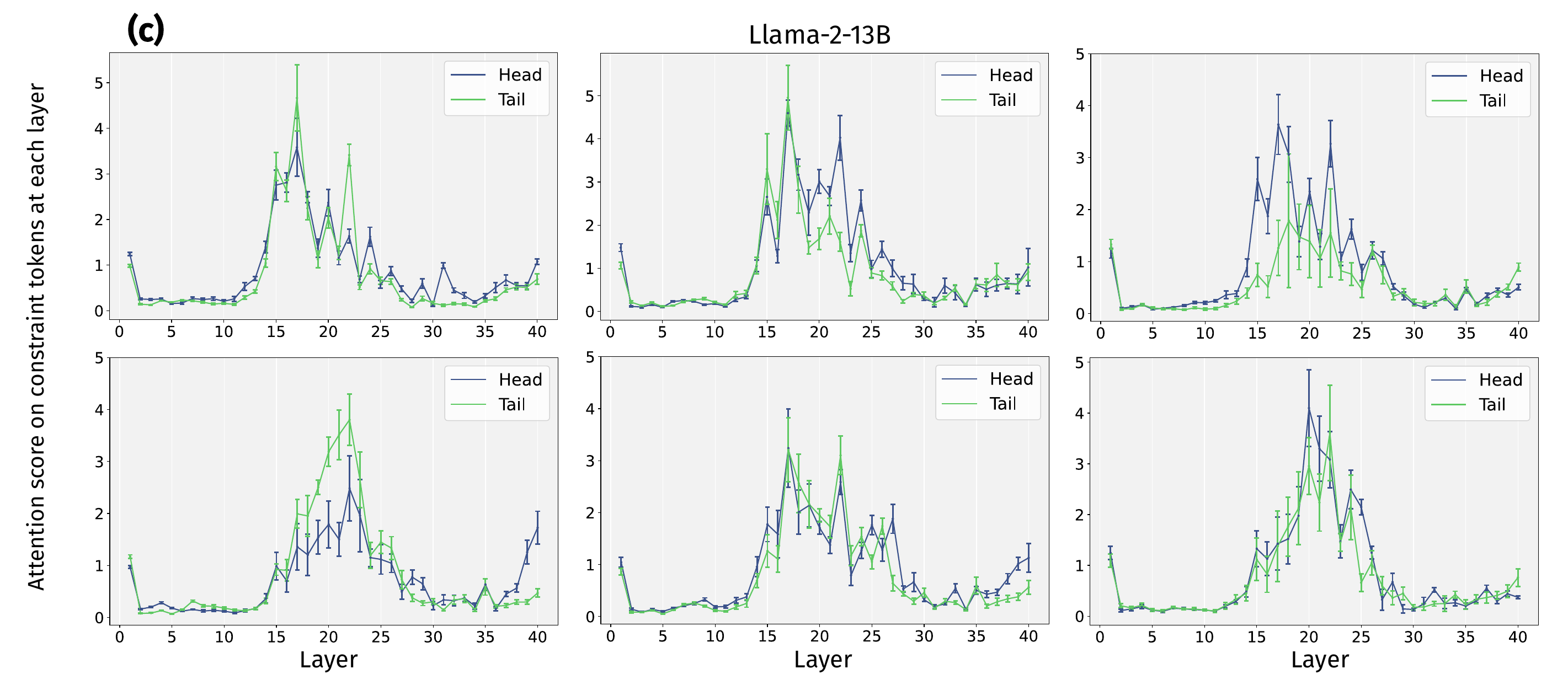} \\
\includegraphics[width = 0.95\linewidth]{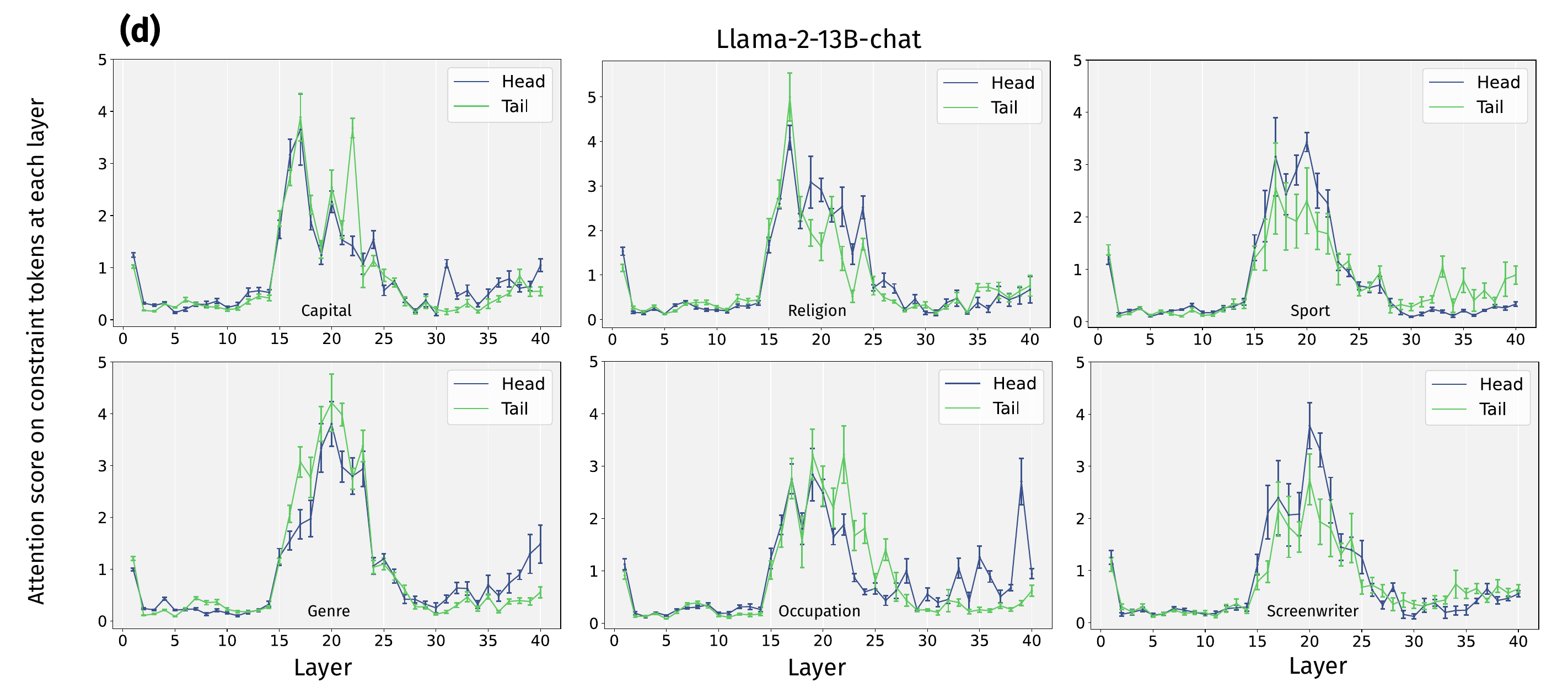} \\
\includegraphics[width = 0.95\linewidth]{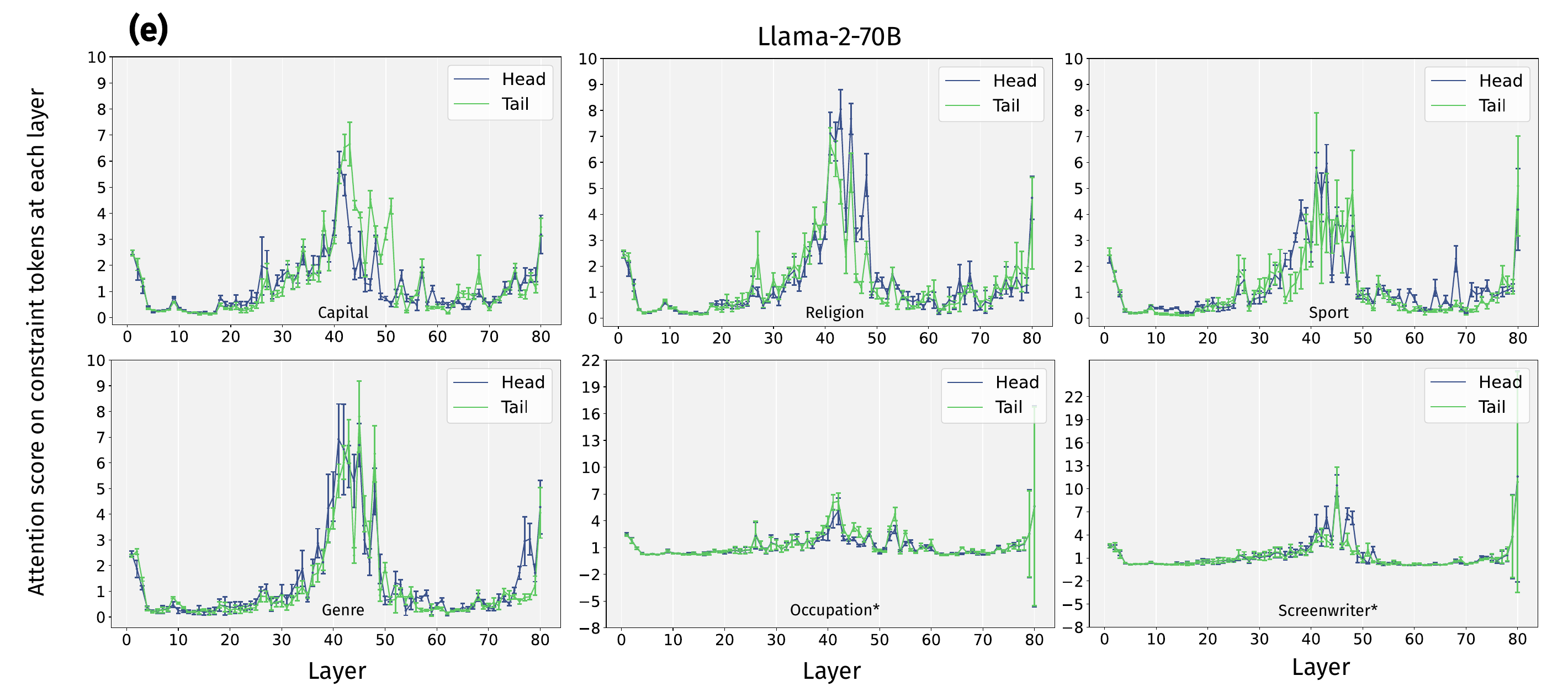} \\
\caption*{\textbf{Figure S2.} \textbf{Effect of popularity on LLM reasoning behaviour}. For each lexical variant of a question, we calculate a measure
of attention scores over each constrain token. For each question, we take a mean of this attention score measure across all its
lexical variants and show it with solid lines for each layer under study. We also take the standard deviation across all of the
lexical variants of a question and depict it using error bars for each layer. {\tt Head} denotes the batch of highly popular questions
while {\tt Tail} denotes the batch of lowly popular questions. The attention score, as well as its sensitivity to lexical variation
(observed from the standard deviation), appears to be highest in the middle layers of processing. Furthermore, we observe that
between {\tt Head} and {\tt Tail}, a lower attention score is associated with the batch of highly popular questions. (Note, Llama-2 70B + the relations `Occupation' and `Screenwriter', showed some unexpected results in the final layers where attention scores were accumulated over the default value of $-1$ used in code (to facilitate obtaining the maximum)).}
\end{longtable}

\subsection*{S2.3 Effect of Popularity in LLM Fact Retrieval Behaviour}
We provide results across all models and relations for assessing the effect of popularity on the internal fact retrieval mechanisms of LLMs.
\begin{longtable}{c} 
\includegraphics[width = 0.95\linewidth]{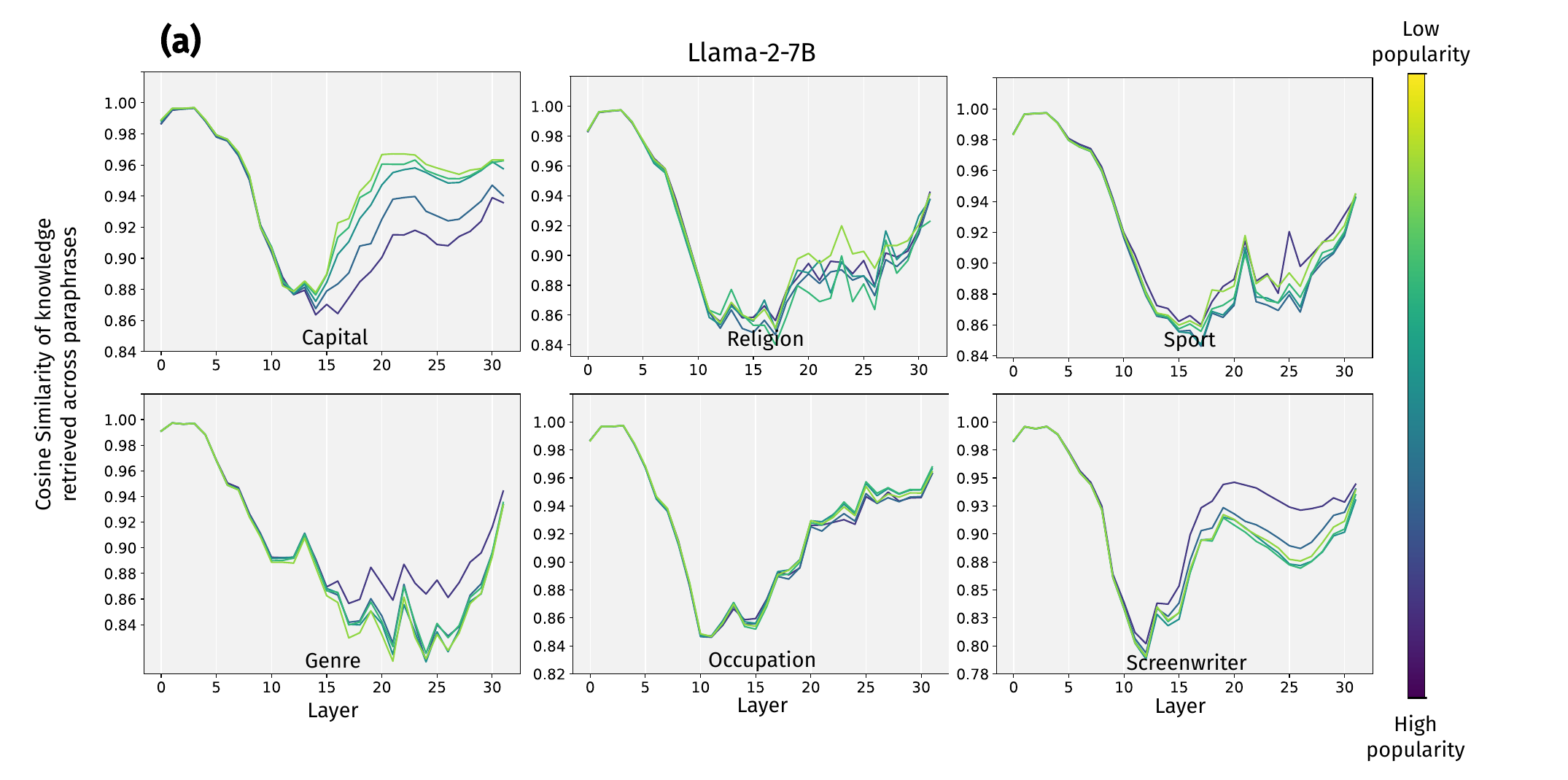} \\
\includegraphics[width = 0.95\linewidth]{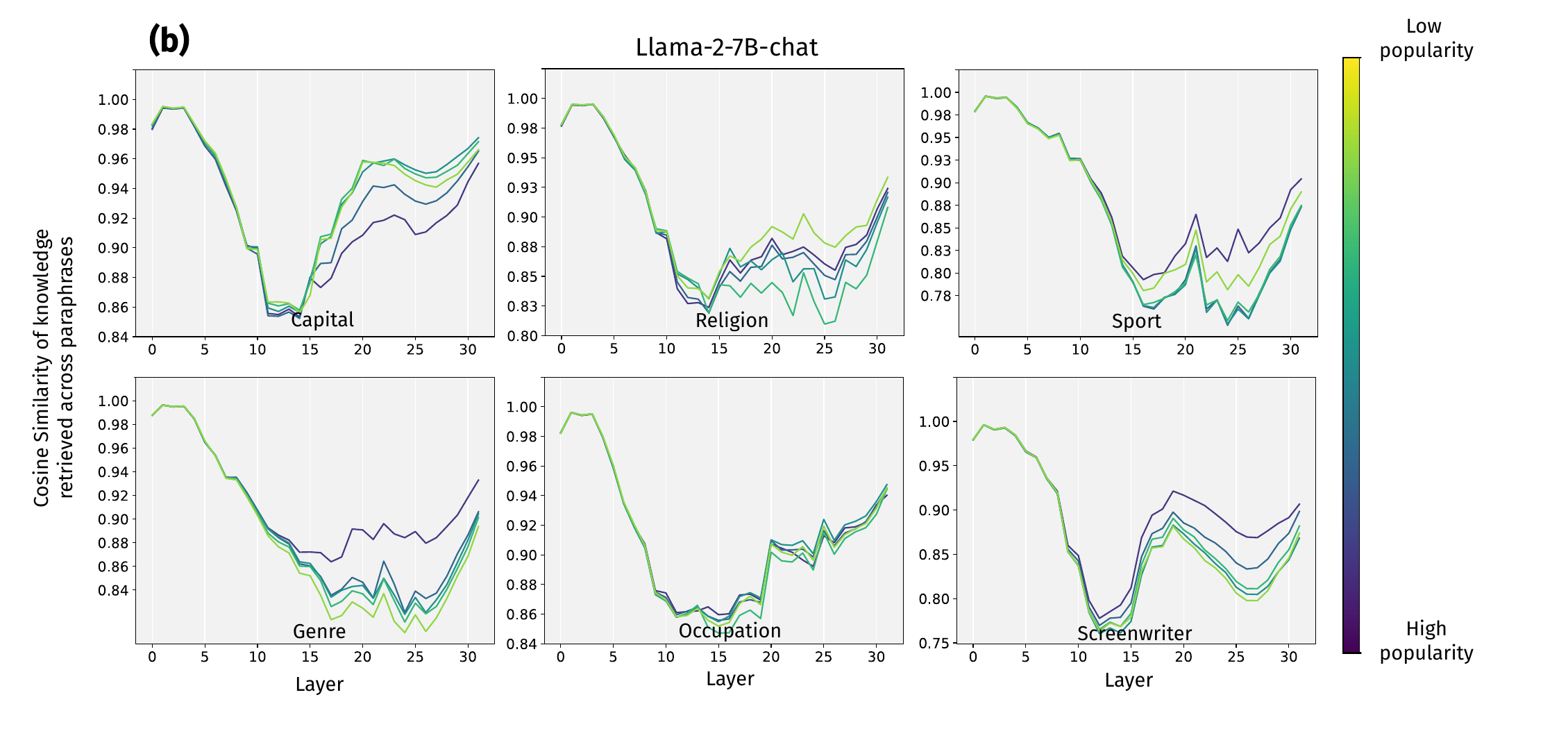} \\
\includegraphics[width = 0.95\linewidth]{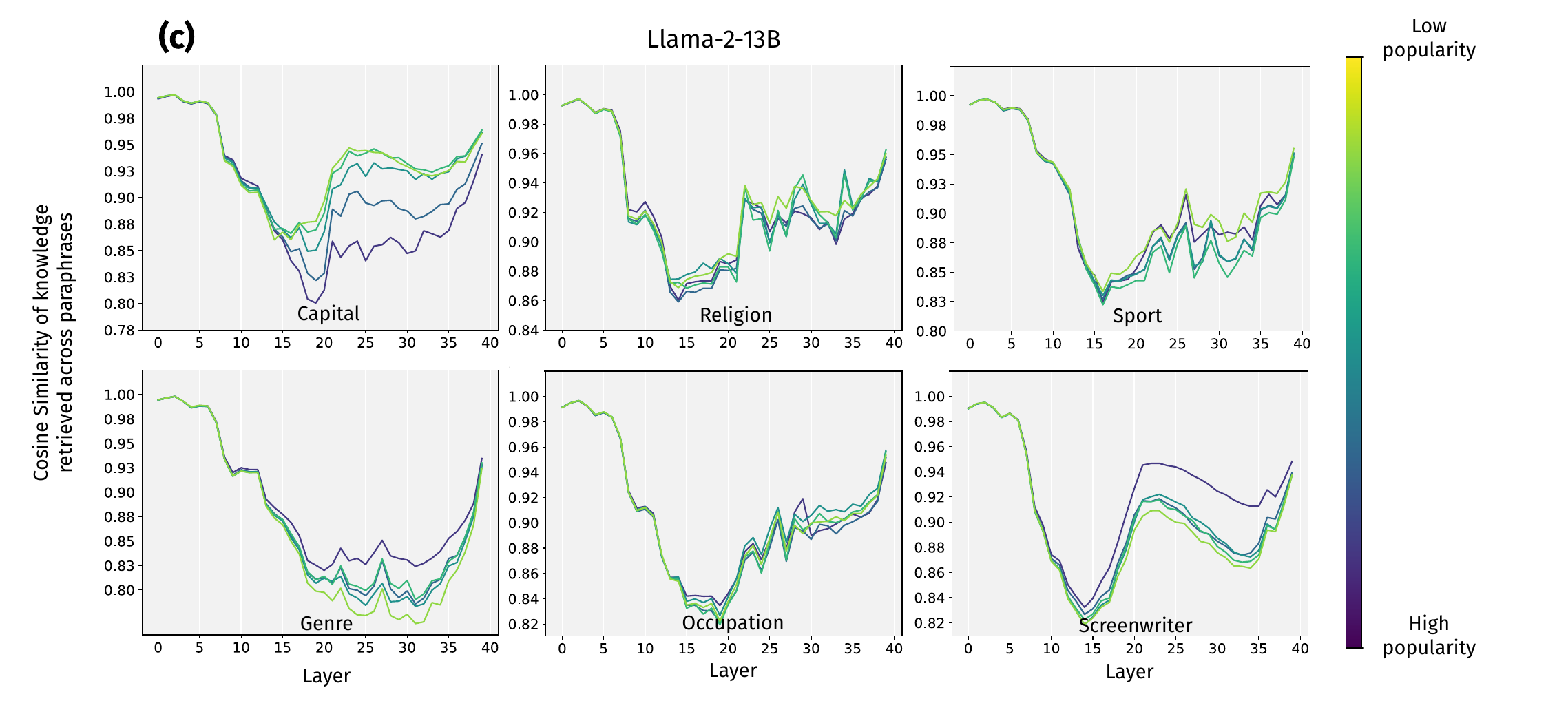} \\
\includegraphics[width = 0.95\linewidth]{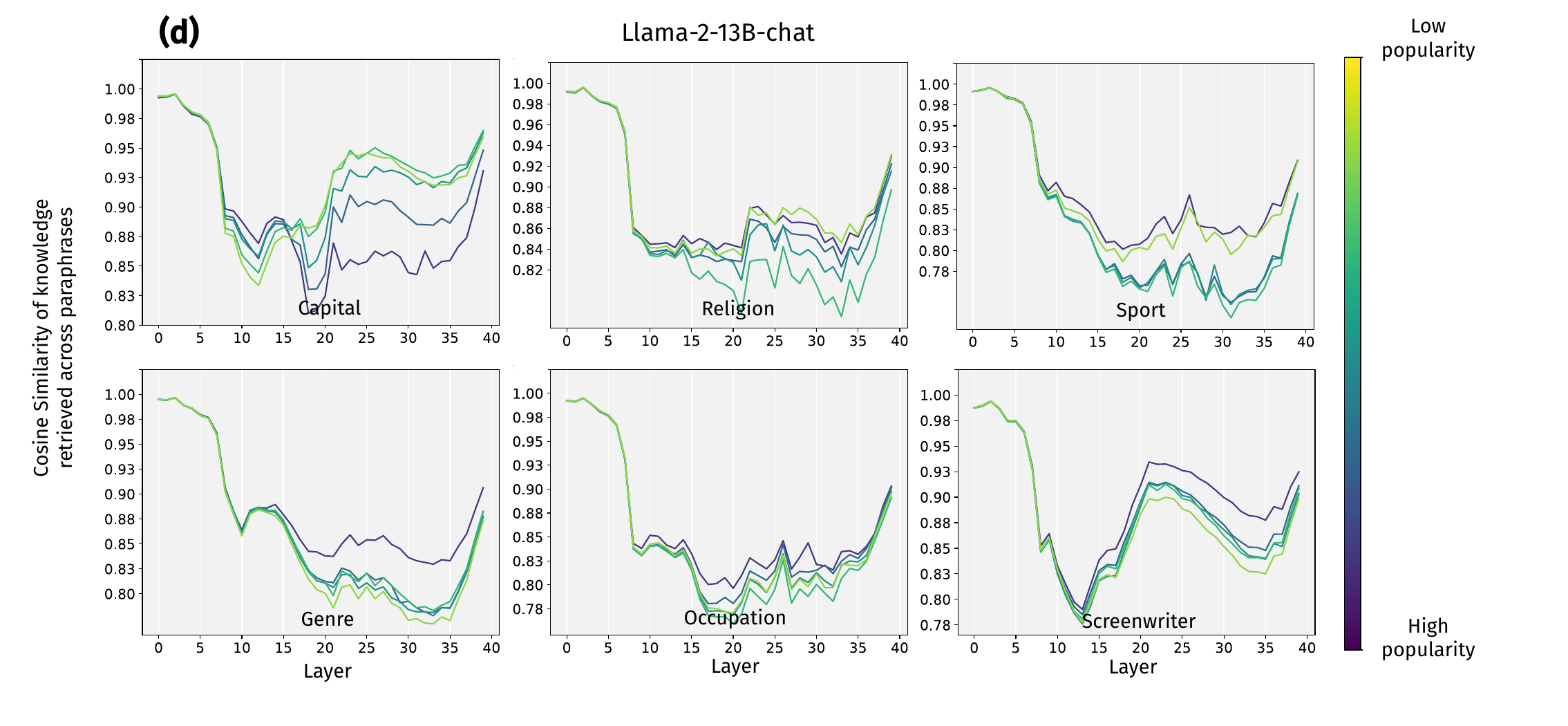} \\
\includegraphics[width = 0.95\linewidth]{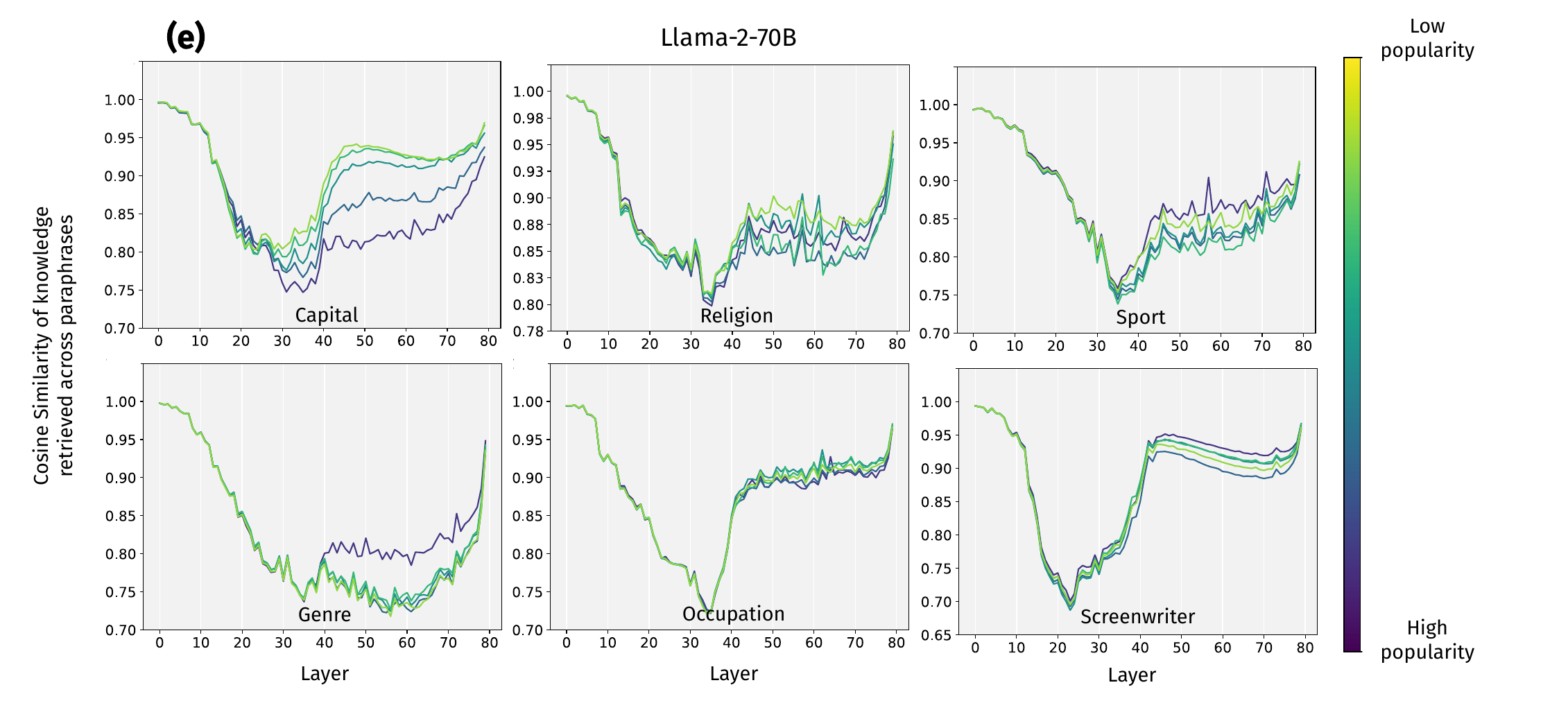} \\
\caption*{\textbf{Figure S3.} \textbf{Effect of popularity on LLM fact-retrieval behaviour}. The figure demonstrates the mean of the cosine similarity of the upward projections intercepted for each of the lexical variations of a question. We observe a lower similarity between the knowledge retrieved across the lexical variations of a highly popular question.}
\end{longtable}

\subsection*{S2.4 Similarity of Knowledge Retrieved among Relations}
We provide results across all models for assessing if LLMs can suitably segregate information about highly popular subjects belonging to distinct relations.
\begin{longtable}{c} 
\includegraphics[width = 0.95\linewidth]{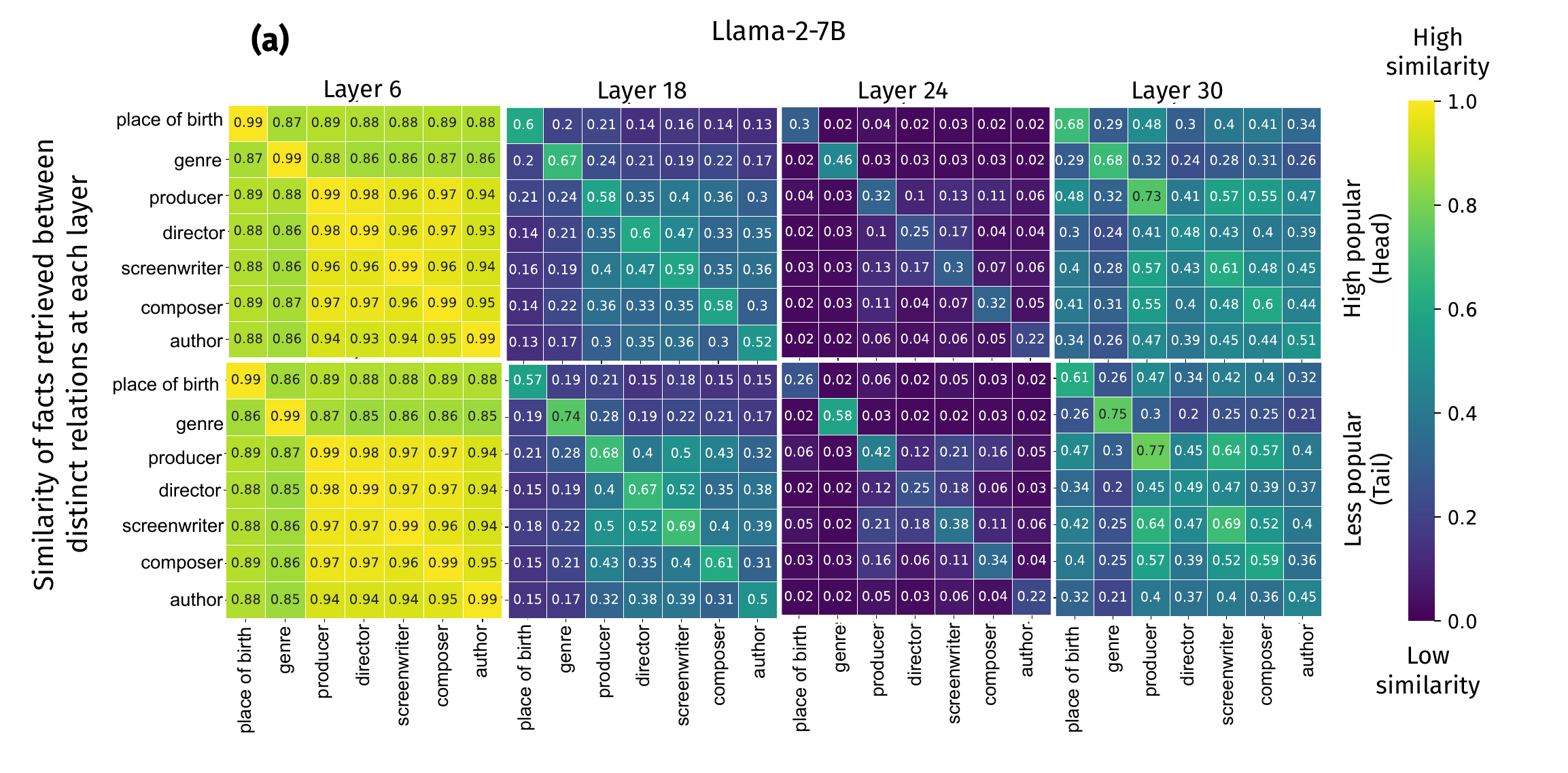} \\
\includegraphics[width = 0.95\linewidth]{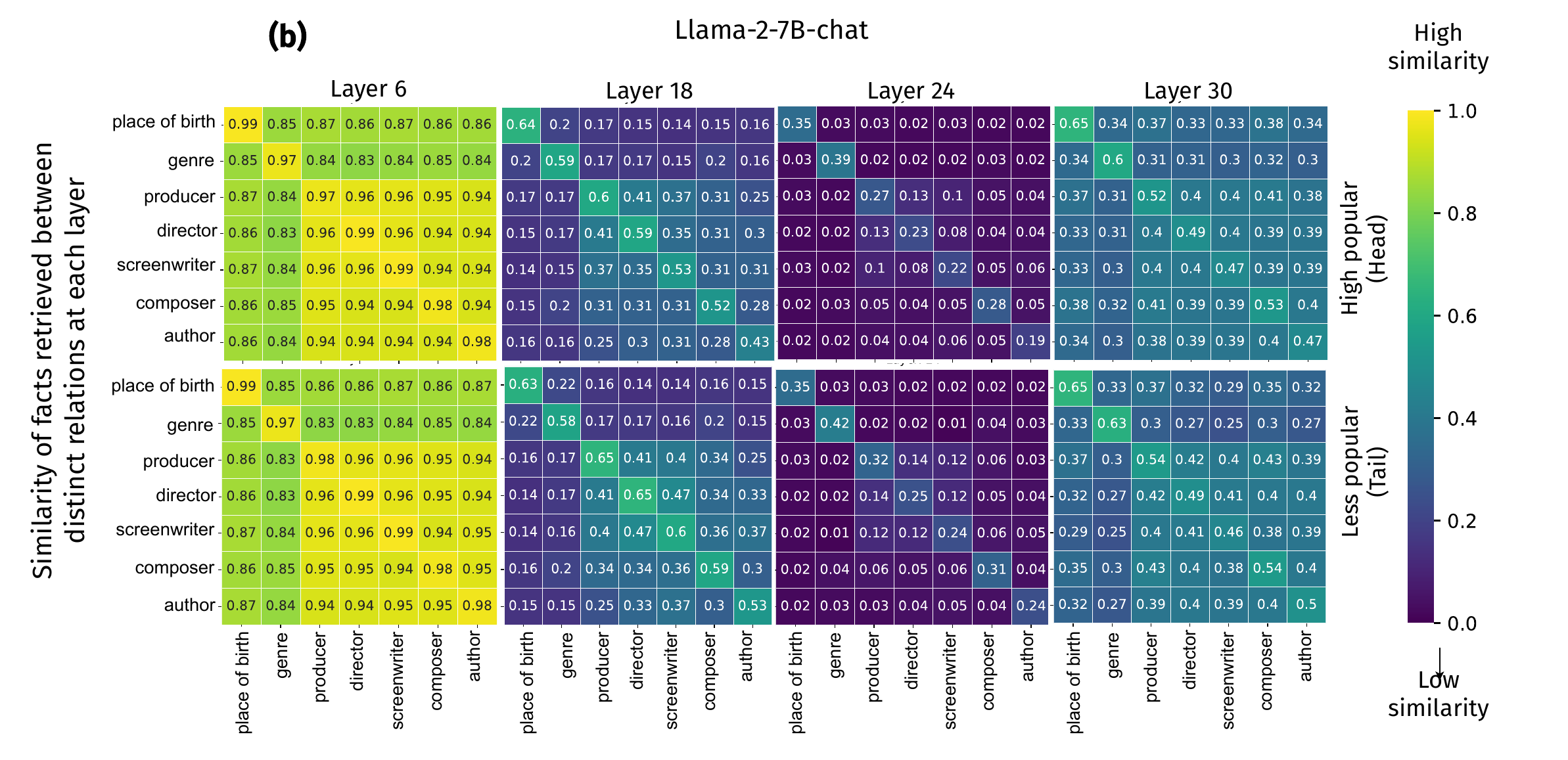} \\
\includegraphics[width = 0.95\linewidth]{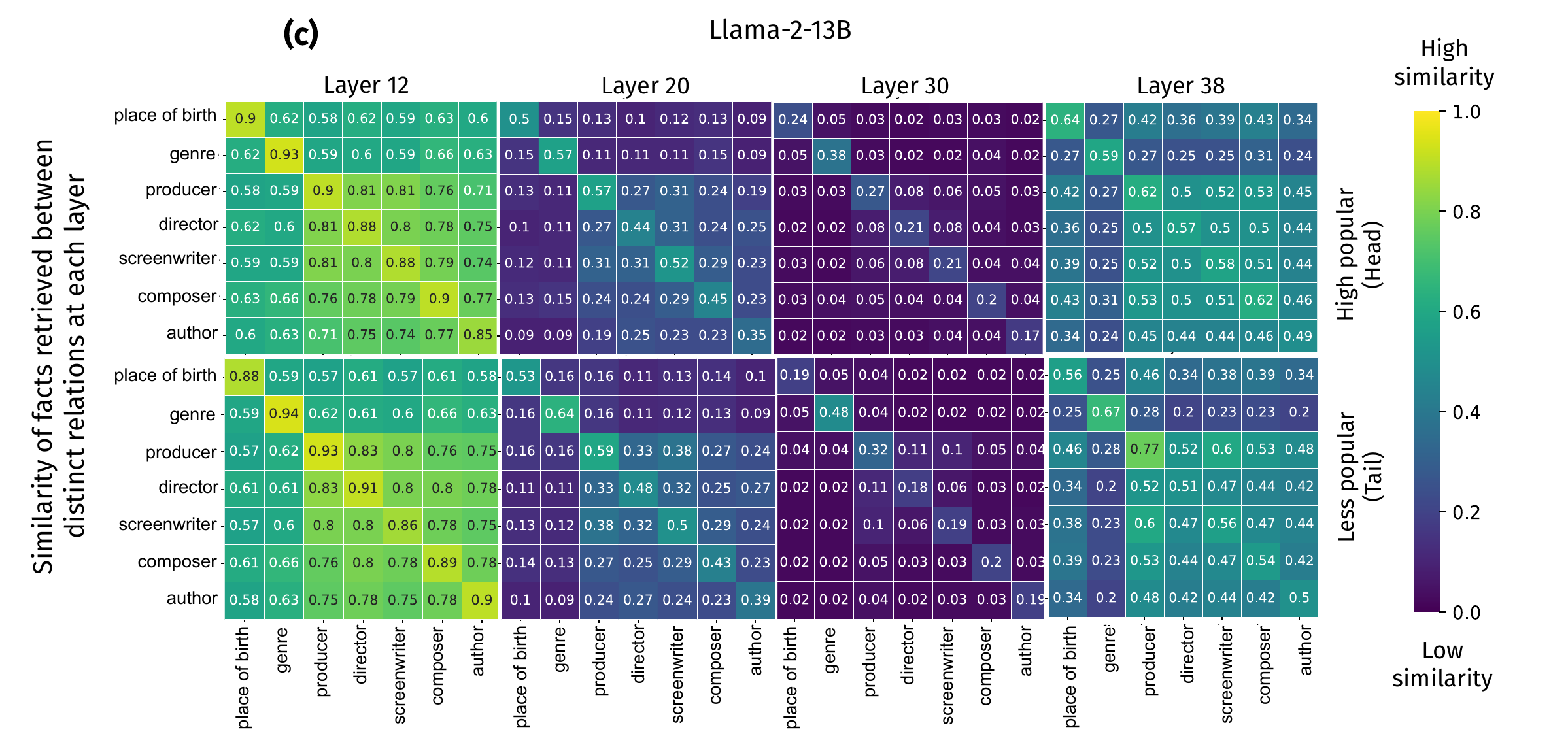} \\
\includegraphics[width = 0.95\linewidth]{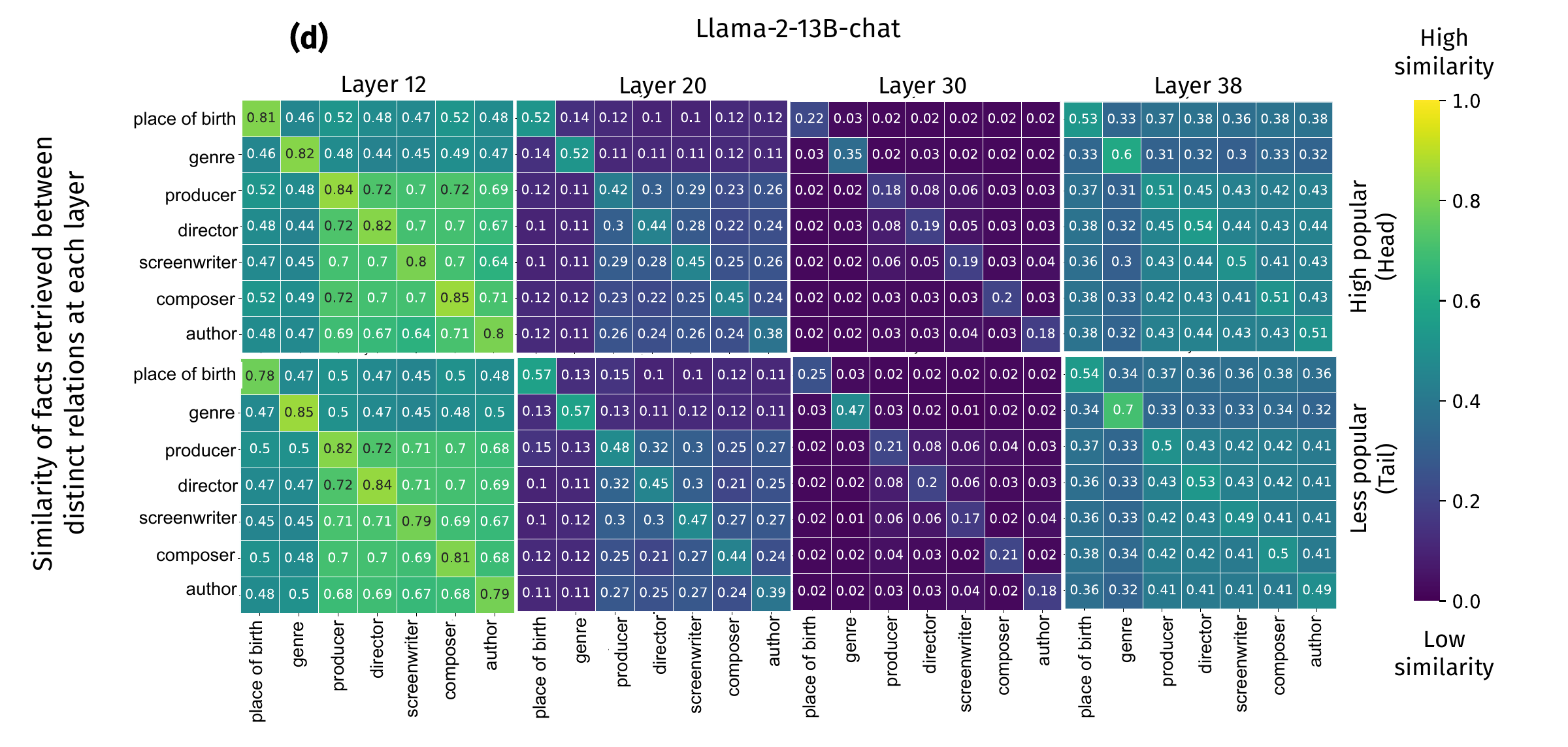} \\
\includegraphics[width = 0.95\linewidth]{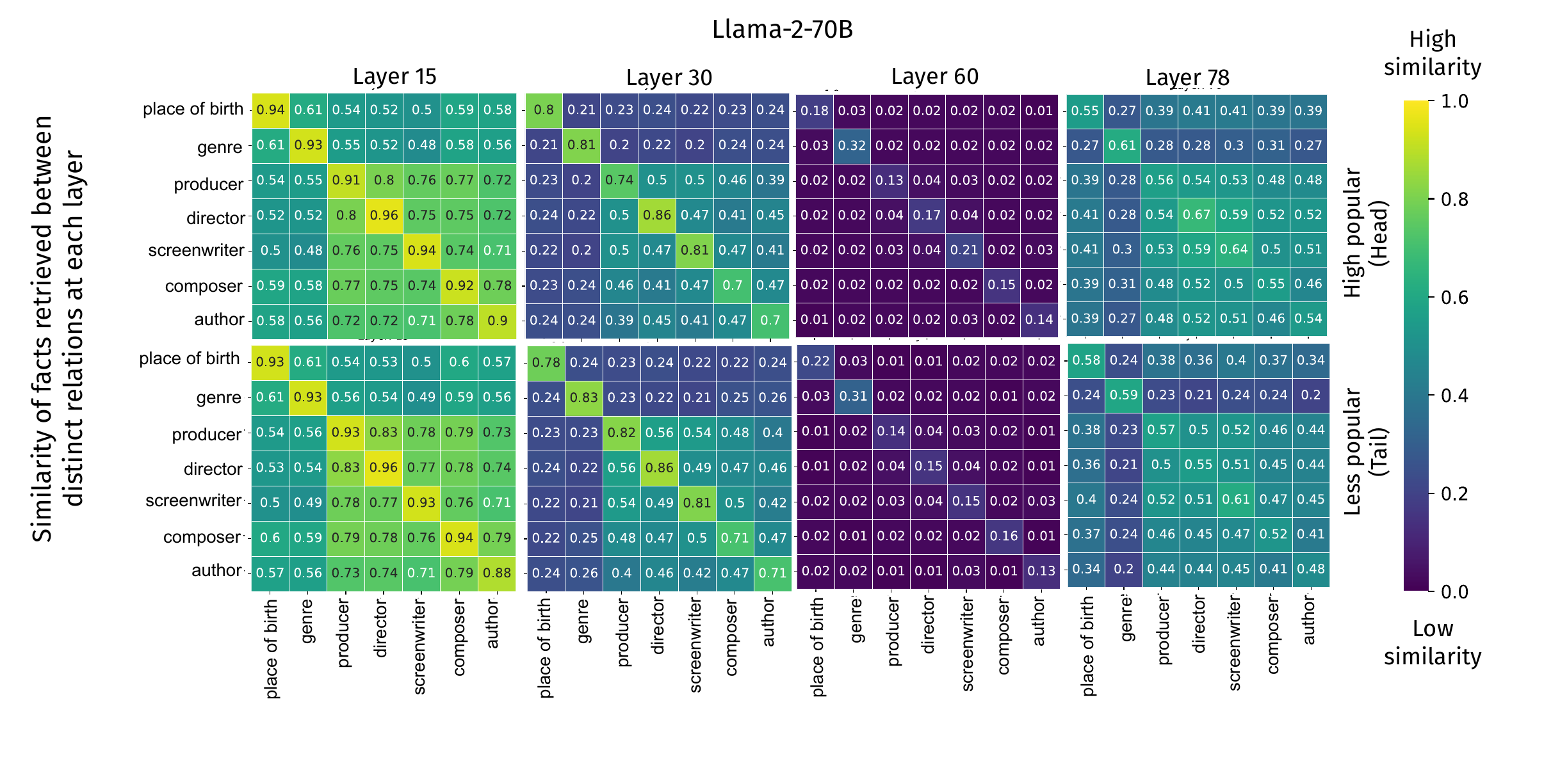} \\
\caption*{\textbf{Figure S4.} \textbf{Similarity of facts retrieved among relations}. For each pair of relations, we show the mean of the similarity of
facts retrieved across all questions at each layer. For both the highly and less popular batches of questions, the similarity of the
knowledge retrieved between any two pairs of relations first decreases across the layers and then increases in the final layers of
processing. For any pair of relations, we observe that there is a higher similarity between the facts retrieved across almost all
layers for highly popular questions.}
\end{longtable}

\end{document}